\documentclass[10pt,journal,compsoc]{IEEEtran}

\usepackage[utf8]{inputenc} % allow utf-8 input
\usepackage[T1]{fontenc}    % use 8-bit T1 fonts
\usepackage{url}            % simple URL typesetting
\usepackage{booktabs}       % professional-quality tables
\usepackage{amsfonts}       % blackboard math symbols
\usepackage{amsmath}
\usepackage{nicefrac}       % compact symbols for 1/2, etc.
\usepackage{microtype}      % microtypography
\usepackage{amsthm}
\usepackage{amssymb}
\usepackage{rotating}
\usepackage{booktabs}
\usepackage{graphicx}
\usepackage{subfigure}
\usepackage{caption}
\usepackage{wrapfig}
\usepackage{bm}
\usepackage{color}
\usepackage{xcolor}
\usepackage{multirow}
\usepackage{mathtools}
\usepackage{lscape}
\usepackage{hyperref}
\usepackage{url}
\usepackage{bbm}
\newcommand{\eat}[1]{}

\usepackage{tikz}   %added by chenghong 
\usepackage{svg}
\usepackage{bbding}
\usepackage{cleveref} %added by chenghong, use cref for IEEE template
\usepackage{algorithmic}
\usepackage{algorithm}
\usepackage{multirow}
\usepackage{makecell}
\usepackage{amsmath,amsfonts,bm}

\def\eqref#1{equation~\ref{#1}}
\def\1{\bm{1}}

\DeclareMathAlphabet{\mathsfit}{\encodingdefault}{\sfdefault}{m}{sl}
\SetMathAlphabet{\mathsfit}{bold}{\encodingdefault}{\sfdefault}{bx}{n}

\definecolor{citeblue}{HTML}{0071bc}
\hypersetup{
  colorlinks,
  linkcolor = citeblue,
  citecolor = citeblue,
}

\usepackage{color,colortbl}
\definecolor{ballblue}{rgb}{0.13, 0.67, 0.80}
\definecolor{amaranth}{rgb}{0.90, 0.17, 0.31}
\definecolor{olive}{rgb}{0.5, 0.5, 0.0}
\definecolor{Gray}{gray}{0.85}
\newcolumntype{a}{>{\columncolor{Gray}}c}
\definecolor{blush}{rgb}{0.87, 0.36, 0.51}

\definecolor{darkgreen}{rgb}{0.0, 0.5, 0.0}
\definecolor{darkred}{rgb}{0.55, 0.0, 0.0}   % added by chenghong for successfully compiling 
\definecolor{deeppeach}{rgb}{1.0, 0.8, 0.64}  % added by chenghong
\definecolor{lightgray}{rgb}{0.83, 0.83, 0.83}  % added by chenghong

\usepackage{pifont}% http://ctan.org/pkg/pifont     % added by chenghong
\newcommand{\thename}{MVHumanNet}

\ifCLASSOPTIONcompsoc
  \usepackage[nocompress]{cite}
\else
  \usepackage{cite}
\fi

\ifCLASSINFOpdf
\else
\fi

\begin{document}

\title{MVHumanNet++: A Large-scale Dataset of Multi-view Daily
Dressing Human Captures with Richer Annotations for 3D Human Digitization}
%\title{Graph Neural Networks and Transformer in Computer Vision: A Survey}

\author{Chenghong Li, Hongjie Liao, Yihao Zhi, Xihe Yang, Zhengwentai Sun, Jiahao Chang\\ Shuguang Cui~\IEEEmembership{Fellow,~IEEE} and Xiaoguang Han$^{\dag}$~\IEEEmembership{Member,~IEEE}% <-this % stops a space
	\IEEEcompsocitemizethanks{
		\IEEEcompsocthanksitem C. Li, H. Liao, Y. Zhi, X. Yang, Z. Sun, J. Chang, S. Cui and X. Han are currently with the School of Science and Engineering, The Chinese University of Hong Kong, Shenzhen. C. Li, Y. Zhi, Z. Sun, J. Chang, S. Cui and X. Han are also with the Future Network of Intelligence Institute, CUHK-Shenzhen. (email: \{chenghongli, hongjieliao, yihaozhi1, xiheyang1, zhengwentaisun, jiahaochang\}@link.cuhk.edu.cn, \{shuguangcui, hanxiaoguang\}@cuhk.edu.cn).

        % \IEEEcompsocthanksitem C. Chen, Y. Wu, Q. Dai, and H.-Y. Zhou contributed equally to this work. Corresponding authors: S. Yang, X. Han, and Y. Yu.
        \IEEEcompsocthanksitem $\dag$ denotes corresponding author.
        
	}}  %TODO: fill this completely 

\markboth{}{Chen \MakeLowercase{\textit{et al.}}: A Survey on Graph Neural Networks and Graph Transformers in Computer Vision}

\IEEEtitleabstractindextext{%
\begin{abstract}
In this era, the success of large language models and text-to-image models can be attributed to the driving force of large-scale datasets. However, in the realm of 3D vision, while significant progress has been achieved in object-centric tasks through large-scale datasets like Objaverse and MVImgNet, human-centric tasks have seen limited advancement, largely due to the absence of a comparable large-scale human dataset. To bridge this gap, we present \textbf{\thename++}, a dataset that comprises multi-view human action sequences of \textbf{4,500} human identities. The primary focus of our work is on collecting human data that features a large number of diverse identities and everyday clothing using multi-view human capture systems, which facilitates easily scalable data collection. Our dataset contains \textbf{9,000} daily outfits, \textbf{60,000} motion sequences and \textbf{645 million} frames with extensive annotations, including human masks, camera parameters, 2D and 3D keypoints, SMPL/SMPLX parameters, and corresponding textual descriptions.  Additionally, the proposed \thename++ dataset is enhanced with newly processed normal maps and depth maps, significantly expanding its applicability and utility for advanced human-centric research. To explore the potential of our proposed \thename++ dataset in various 2D and 3D visual tasks, we conducted several pilot studies to demonstrate the performance improvements and effective applications enabled by the scale provided by \thename++. As the current largest-scale 3D human dataset, we hope that the release of \thename++ dataset with annotations will foster further innovations in the domain of 3D human-centric tasks at scale. \thename++ is publicly available at 
\url {https://kevinlee09.github.io/research/MVHumanNet++/}.

\end{abstract}

}

\maketitle

\IEEEdisplaynontitleabstractindextext

\IEEEpeerreviewmaketitle
\IEEEraisesectionheading{\section{Introduction}\label{sec:introduction}}
\IEEEPARstart{I}{n} recent years, the exponential advancements of AI have been largely driven by the massive amounts of data. In the field of computer vision, with the emergency of SA-1B~\cite{Kirillov_2023_ICCV} and LAION-5B~\cite{schuhmann2022laion}, models like SAM~\cite{Kirillov_2023_ICCV} and Stable Diffusion~\cite{rombach2022high} have greatly benefited from these large volumes of data, enabling zero-shot transfer to downstream tasks.  Subsequently, Objaverse~\cite{deitke2023objaverse, deitke2023objaversexl} and MVImgNet~\cite{yu2023mvimgnet} break barriers of 3D data collection with large-scale synthetic 3D assets and real-world multi-view capture, which support Zero123~\cite{liu2023zero} and LRM~\cite{hong2023lrm} models to achieve impressive generalization ability of multi-view or 3D reconstruction. However, comparable progress on human-centric tasks still remained elusive due to the limited scale of 3D human data.

\begin{figure*}[!t]
    \centering
    \captionsetup{type=figure}
    \includegraphics[width=1.0\textwidth]{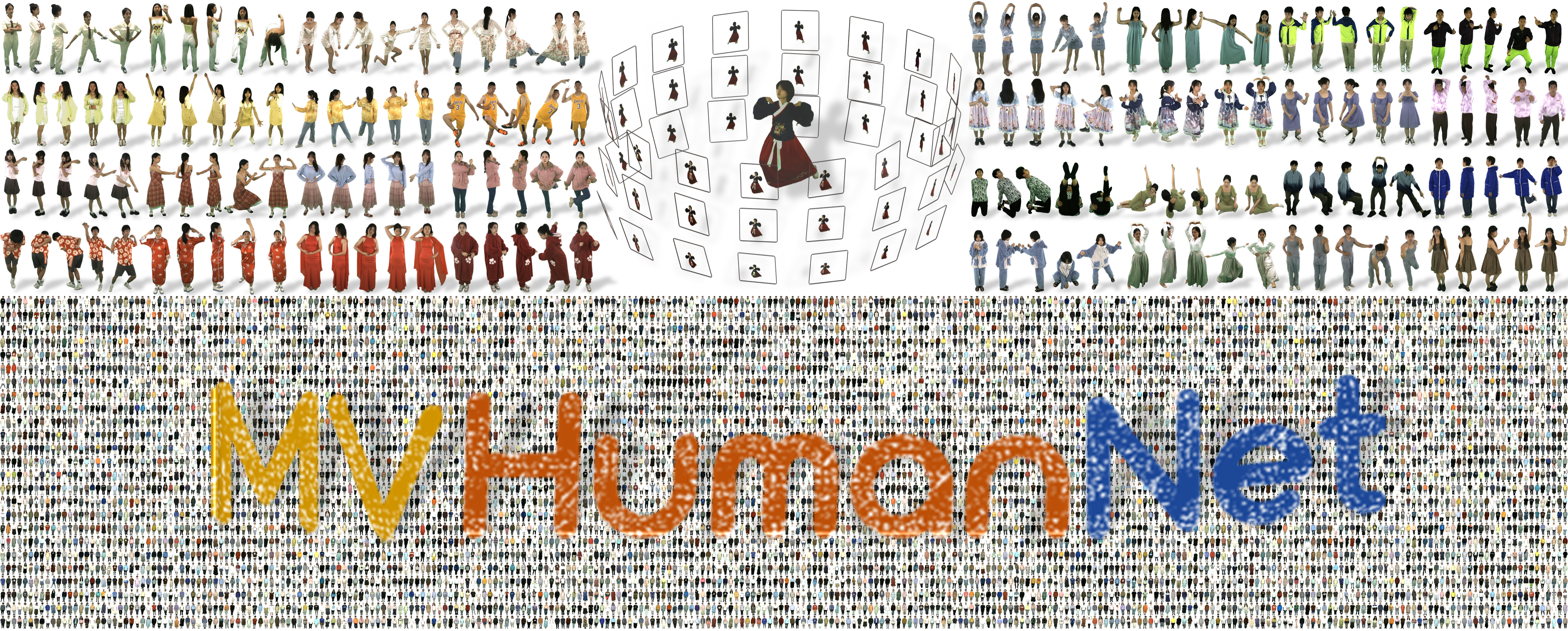}
    \captionsetup{skip=5pt} % 调整所需的间距
    \caption{
    We introduce \textbf{\thename++}, a large-scale dataset of \textit{multi-view human images} with unprecedented scale in human subjects, daily outfits, motion sequences and frames. 
    \textbf{Top left and right}: Examples of multi-view poses featuring different human identities with various daily dressing in our dataset. 
    \textbf{Top middle}: Our multi-view capture system includes 48 cameras of 12MP resolution. 
    \textbf{Bottom}: Comprehensive visualization of all 9000 outfits in our \thename++.
    \label{fig:teaser}}
    \vspace{-3mm}
\end{figure*}

Compared to collecting 3D object datasets, capturing high-quality and large-scale 3D human avatars is more time-consuming in the same order of scale. Existing 3D human datasets can be categorized into two distinct representations: 3D human scans and multi-view human images. While 3D human scan data~\cite{RenderPeople, yu2021function4d} provides accurate geometric shapes, it comes with high acquisition costs which leads to limited data scale. Conversely, multi-view capture provides an easier way to collect 3D human data. Previous multi-view human datasets~\cite{peng2021neural,ionescu2013human3,li2021ai} involve only a few human subjects. Recent advances in multi-view human data~\cite{cheng2022generalizable,cheng2023dna} narrow the gap of data scarcity which provides more representative human data for establishing reasonable benchmarks. To ensure comprehensiveness, it is necessary for these datasets to consider the complex clothing and the uncommon human-object interaction. However, incorporating these factors introduces complexities for scaling up the dataset.

%as well as the different ethnicities, 不想提ethnicities

To scale up the 3D human data, we present \textbf{\thename++}, a large-scale multi-view human performance capture dataset. Our dataset primarily focuses on casual clothing commonly found in everyday life, enabling to easily expand the scale of human data collection. For the hardware setup, we establish two 360-degree indoor systems equipped with 48 and 24 calibrated RGB cameras, respectively, to capture high-fidelity videos with resolutions up to 12MP (4096 × 3000)  and 5MP (2048 × 2448). Considering the capture of human data, we intend to cover a wide range of attributes among human subjects, including age, body shape, motion, as well as the colors, types, and materials of dressing, enabling our dataset as diverse as possible. Furthermore, we also design 500 motion types to guarantee coverage of daily scenarios. Overall, we invite \textbf{4,500} individuals to participate in data capture process. Each participant is recorded in two distinctive outfits (\textbf{9,000} in total) and seven different motion sequences. Thanks to the targeted collection of everyday clothing,  data capture for each participant has been accomplished efficiently within six months. Eventually, the full dataset comprises an extensive collection of \textbf{60,000} motion sequences with over \textbf{645 million} frames. Compared with the existing multi-view human datasets~\cite{ionescu2013human3, peng2021neural, cheng2022generalizable, isik2023humanrf}, \thename++ provides a significantly larger number of human subjects and outfits than previously available. Furthermore, \thename++ surpasses the recently proposed DNA-Rendering~\cite{cheng2023dna} dataset by an order of magnitude in terms of motion and frame data.
The detailed comparisons between \thename++  and other relevant datasets are shown in Table. \ref{tab:dataset_comparison}.

\begin{table*}[htbp]
    \centering
    \resizebox{1.0\textwidth}{!}{%
    \begin{tabular}{c|ccc|ccccc|c}
        \toprule
        Dataset & Age & Cloth & Motion & \#ID & \#Outfit & \#Actions & \#View & \#Frames & Resolution\\
        \midrule
        Human3.6M~\cite{ionescu2013human3} & \textcolor{darkred}{\XSolidBrush} & \textcolor{darkred}{\XSolidBrush} & \textcolor{darkgreen}{\textcolor{darkgreen}{\CheckmarkBold}} & 11  &  11 & 17 & 4 & 3.6M & 1000P \\
        CMU Panoptic~\cite{joo2015panoptic} & \textcolor{darkgreen}{\CheckmarkBold} & \textcolor{darkred}{\XSolidBrush} & \textcolor{darkgreen}{\CheckmarkBold} & 97  & 97 & 65 & 31 & 15.3M & 1080P \\
       MPI-INF-3DHP~\cite{mehta2017monocular} & \textcolor{darkred}{\XSolidBrush} & \textcolor{darkred}{\XSolidBrush} & \textcolor{darkgreen}{\CheckmarkBold} & 8  & 8 & $-$  & 14 &  1.3M & \cellcolor{lightgray}2048P \\
        NHR~\cite{wu2020multi} & \textcolor{darkred}{\XSolidBrush} & \textcolor{darkred}{\XSolidBrush} & \textcolor{darkgreen}{\CheckmarkBold} & 3  & 3 & 5 & 80 & 100K & \cellcolor{lightgray}2048P \\
        ZJU-MoCap~\cite{peng2021neural} & \textcolor{darkred}{\XSolidBrush} & \textcolor{darkred}{\XSolidBrush} & \textcolor{darkgreen}{\CheckmarkBold} & 10  & 10 & 10 & 24 & 180K & 1024P \\
        Neural Actor~\cite{liu2021neural} & \textcolor{darkred}{\XSolidBrush} & \textcolor{darkred}{\XSolidBrush} & \textcolor{darkgreen}{\CheckmarkBold} & 8  & 8 & $-$  & 11$\sim$100 & 250K & 1285P \\
        HUMBI~\cite{yu2020humbi} & \textcolor{darkgreen}{\CheckmarkBold} & \textcolor{darkgreen}{\CheckmarkBold} & \textcolor{darkred}{\XSolidBrush} & 772  & 772  & $-$ & \cellcolor{lightgray}107 & 26M & 1080P \\
        AIST++~\cite{li2021ai} & \textcolor{darkred}{\XSolidBrush} & \textcolor{darkred}{\XSolidBrush} & \textcolor{darkred}{\XSolidBrush} & 30  & 30 & $-$ & 9 & 10.1M & 1080P \\
        THuman 4.0~\cite{zheng2022structured} & \textcolor{darkred}{\XSolidBrush} & \textcolor{darkred}{\XSolidBrush} & \textcolor{darkgreen}{\CheckmarkBold} & 3  & 3 & $-$ & 24 & 10K & 1150P \\
        HuMMan~\cite{cai2022humman} & \textcolor{darkred}{\XSolidBrush} & \textcolor{darkgreen}{\CheckmarkBold} & \textcolor{darkgreen}{\CheckmarkBold} & \cellcolor{lightgray}1000  & 1000 & \cellcolor{lightgray}500 & 10 & 60M & 1080P \\
        GeneBody~\cite{cheng2022generalizable} & \textcolor{darkgreen}{\CheckmarkBold} & \textcolor{darkgreen}{\CheckmarkBold} & \textcolor{darkgreen}{\CheckmarkBold} & 50  & 100 & 61 & 48 & 2.95M & \cellcolor{lightgray}2048P \\
        ActorsHQ~\cite{isik2023humanrf} & \textcolor{darkred}{\XSolidBrush} & \textcolor{darkred}{\XSolidBrush} & \textcolor{darkgreen}{\CheckmarkBold} & 8  & 8 & 52 & \cellcolor{deeppeach}160 & 40K & \cellcolor{deeppeach}4096P \\
        DNA-Rendering~\cite{cheng2023dna} & \textcolor{darkgreen}{\CheckmarkBold} & \textcolor{darkgreen}{\CheckmarkBold} & \textcolor{darkgreen}{\CheckmarkBold} & \cellcolor{lightgray}500  & \cellcolor{lightgray}1500 & \cellcolor{deeppeach}1187 & 60 & \cellcolor{lightgray}67.5M & \cellcolor{deeppeach}4096P \\
        \midrule
        \textit{\textbf{MVHumanNet++(Ours)}} & \textcolor{darkgreen}{\CheckmarkBold} & \textcolor{darkgreen}{\CheckmarkBold} & \textcolor{darkgreen}{\CheckmarkBold} & \cellcolor{deeppeach}4500  & \cellcolor{deeppeach}9000 & \cellcolor{lightgray}500 & 48 & \cellcolor{deeppeach}645.1M & \cellcolor{deeppeach}4096P \\
        \bottomrule
    \end{tabular}%
    }
        \captionsetup{skip=6pt} % Adjust the value as desired
     \caption{\textbf{Dataset comparison on existing multi-view human-centric datasets.} \thename++ provides a significantly larger number of human subjects and outfits than previous datasets available, regarding the number of identities (\#ID), outfits in total (\#Outfit) and frames of images (\#Frames). Attributes among humans, including age, cloth and motion are covered (denoted by \textcolor{darkgreen}{\text{\ding{51}}} for inclusion and \textcolor{darkred}{\text{\ding{55}}} for exclusion.). Cells highlighted in {\begin{tikzpicture} [scale=1.0] \fill[deeppeach] (0,0) rectangle (1em,1em); \end{tikzpicture}}   {\begin{tikzpicture}[scale=1.0]  \fill[color=lightgray] (0ex,0ex) rectangle (1em,1em);  \end{tikzpicture} denotes the dataset with the best and second-best feature in each column.}
     \vspace{-2mm}
}
     % \caption{\textbf{Dataset comparison on diversity~\lhjnote{diversity?} and scales.} MVHumanNet provides a significantly larger number of human subjects and outfits than previous datasets available, regarding the number of identities (\#ID), outfits in total (\#Outfit) and frames of images (\#Frames). Attributes among humans, including age, cloth and motion are covered  (which are marked with \textcolor{darkgreen}{\text{\ding{51}}} and \textcolor{darkred}{\text{\ding{55}}}). Note that superscript \textsuperscript{*} means low-resolution VGA cameras, which are excluded during "\#Frame" calculation. Best and second best are highlighted in each column~\lhjnote{need to distinguish from dna-rendering} with cells. The best and second-best in each column are highlighted with {\begin{tikzpicture} [scale=1.0] \fill[deeppeach] (0,0) rectangle (1em,1em); \end{tikzpicture}}   {\begin{tikzpicture}[scale=1.0]  \fill[color=lightgray] (0ex,0ex) rectangle (1em,1em);  \end{tikzpicture}.}
    \vspace{-2.5mm}
    \label{tab:dataset_comparison}
\end{table*}
In order to benefit downstream human-centric tasks, we provide essential annotations, including action labels, camera intrinsics and extrinsics, human masks, 2D/3D keypoints, SMPL/SMPLX~\cite{loper2023smpl, pavlakos2019smplx} parameters, and text descriptions, complemented by newly processed normal maps and depth maps, to further enhance the applicability of our dataset. To thoroughly explore the capabilities of our dataset, we carefully design several pilot experiments: \textbf{a)} view-consistent action recognition, \textbf{b)} NeRF~\cite{mildenhall2020nerf} reconstruction for human, \textbf{c)} 3D Gaussian Splatting (3DGS)~\cite{kerbl20233d} reconstruction for human, \textbf{d)} Text-driven view-unconstrained human image generation,  \textbf{e)} 2D view-unconstrained human image and 3D avatar generation, along with the synthesis of multi-view human images and \textbf{f)} Fine-tune DUSt3R~\cite{wang2024dust3r} for unconstraint human  reconstruction. First, by leveraging the multi-view nature of human capture data, we can achieve more accurate view-consistent action recognition and enhance the generalization capabilities of NeRF and 3DGS as the data scale increases. Furthermore, the unprecedented scale of subjects, outfits, pose sequences, and paired textual descriptions enables us to fine-tune a remarkable text-driven, pose-conditioned high-quality human image generation model. Additionally, by exploiting large-scale multi-view human images, we can develop 2D/3D or multi-view full-body human generative models with promising results. Finally, we explore the potential of fine-tuning DUSt3R for human reconstruction with unconstrained human images as input. The aforementioned experiments reveal the promise and opportunities with the large-scale \thename++ dataset to boost a wide range of digital human applications and inspire future research.

In summary, the main contributions of our work include: \textbf{1)} We present the largest multi-view human capture dataset, which is nearly ten times larger than DNA-Rendering dataset in terms of human subjects, motion sequences, and frames with more comprehensive annotations. \textbf{2)} We conduct several pilot studies that demonstrate the proposed dataset can support various downstream human-centric tasks.
\textbf{3)} We believe that \thename++ opens up new possibilities for research in the field of 3D human digitization.

This paper extends our conference paper published in CVPR 2024~\cite{xiong2024mvhumannet}. In this version, \textbf{1)} We enhance the quality of masks and SMPL/SMPLX parameters~({Sec. \ref{data_ann}}), which significantly improves the fidelity of human reconstruction~(Sec. \ref{per_subject_gs}). \textbf{2)} We process normal maps and depth maps as prior data~(Sec. \ref{data_ann}) to facilitate advanced human reconstruction tasks~(Sec. \ref{general_gs}). \textbf{3)} We conduct more comprehensive pilot experiments to validate the proposed \thename++’s value in the task of 3D human reconstruction, which improves the performance of human reconstruction models as the data scale increases~(Sec. \ref{avatar_task} and~Sec. \ref{dust3r_task}).

\section{Related Work}
\label{sec:related_work}
\textbf{3D Human Reconstruction and Generation.} 
Recently, we have witnessed impressive performance in the field of image generation, 3D reconstruction and novel view synthesis in computer vision community with the emergency of Generative Adversarial Networks (GANs)~\cite{goodfellow2014gan, isola2017image, karras2019style}, Neural Implicit Function~\cite{park2019deepsdf, mescheder2019occupancy, chen2019learning} and Neural Radiance Field (NeRF)~\cite{mildenhall2020nerf, muller2022instant}. These successes inspire subsequent works~\cite{fruhstuck2022insetgan, saito2019pifu, peng2021neural, kwon2021neural}  to extend reconstruction and generation tasks to high-fidelity clothed full-body humans.
Many efforts have also been made to combine 2D GANs with NeRF representations for 3D-aware, photo-realistic image synthesis. EG3D~\cite{chan2022efficient} proposes the 3D-aware generation of multi-view face images by introducing an efficient tri-plane representation for volumetric rendering.   GET3D~\cite{gao2022get3d} utilizes two separate latent codes to generate the SDF and texture field, enabling the generation of textured 3D meshes. EVA3D~\cite{hong2022eva3d} extends EG3D to learn generative models with human body priors for 3D full-body human generation from a collection of 2D images. HumanGen~\cite{jiang2023humangen} and Get3DHuman~\cite{xiong2023Get3DHuman} further incorporate the priors of StyleGAN-Human~\cite{fu2022styleganhuman} and PIFuHD~\cite{saito2020pifuhd} for generative human model construction. In addition, Text2Human~\cite{jiang2022text2human} and AvatarClip~\cite{hong2022avatarclip} explore to leverage the powerful vision-language model CLIP~\cite{radford2021learning} for text-driven 2D and 3D human generation.  However, these works can only utilize limited real-world human data, which consequently affects the generalizability of their models. Moreover, the current methods of human generation often train their models on datasets comprising only front-view 2D human images~\cite{liu2016deepfashion, fu2022styleganhuman} or monocular human videos~\cite{zablotskaia2019dwnet}. Unfortunately, these approaches fail to produce satisfactory results when altering the input image across various camera viewpoints. In this work, we provide the current largest scale of  multi-view human capture images along with text descriptions to facilitate 3D human-centric tasks.

\noindent \textbf{3D Human Gaussian Splatting.} Recently, 3D Gaussian splatting~\cite{kerbl20233d} has emerged as an alternative 3D representation to NeRF~\cite{mildenhall2020nerf} due to the impressive quality and speed. Some concurrent works utilize human template as the 3D prior and  bind 3D gaussian primitives on the template mesh to create animatable representations~\cite{hu2024gaussianavatar, hu2024gauhuman, lei2024gart, li2024animatable, pang2024ash}. However, these methods are not generalizable and require new optimization process for every new subject. GPS-Gaussian~\cite{zheng2024gps} achieve generalization to novel humans by incorporating a stereo-depth estimation module, which serves as a partial geometry prior. However, they suffer when given sparse views with few overlappings and thus depth could not be estimated.  GHG~\cite{kwon2025generalizable} achieves real-time 3D Gaussian-based human novel view synthesis in a feed-forward manner, but it requires additional human template priors. EVA-Gaussian~\cite{hu2024eva} introduce an efficient cross-View attention module to accurately estimate the depth map from the source images and then integrate the source images with the estimated depth map to predict the attributes and feature embeddings of the 3D Gaussians.

\noindent \textbf{3D Human Scanning Datasets.} Understanding human actions and reconstructing detailed body geometries with realistic appearances are challenging tasks that require high-quality and large-scale human data. Early works~\cite{bogo2014faust, bogo2017dynamic, zhang2017detailed} in this field provide dynamic human scans but with limited data consisting of only a few subjects or simple postures. Parallel works such as Northwestern-UCLA~\cite{wang2014cross} and NTU RGB+D series~\cite{liu2019ntu, shahroudy2016ntu} utilize more affordable Kinect sensors to obtain depth and human skeleton data, enabling the capture of both appearance and action cues. However, due to the limitations in the accuracy of Kinect sensors, these datasets are inadequate for precise human body modeling. Subsequently, AMASS~\cite{AMASS:ICCV:2019} further integrates traditional motion capture datasets~\cite{CMU_mocap, SFU_mocap} and expands them with fully rigged 3D meshes to facilitate advancements in human action analysis and body modeling research. With the emergency of learning-based digital human techniques, relevant algorithms~\cite{saito2019pifu, xiu2022icon, saito2020pifuhd, chen2021snarf} heavily rely on human scan datasets with high-fidelity 3D geometry and corresponding images. Several studies~\cite{zheng2019deephuman, yu2021function4d, zheng2021deepmulticap, shen2023x, han2023high, ma2020learning} capture their own datasets and release the data to the public for research purposes. Additionally, there are several commercial scan datasets~\cite{RenderPeople, Twindom, AXYZ, 3dpeople} that are well-polished and used for research to ensure professional quality. These datasets play a foundational role in bridging the gap between synthetic virtual avatars and real humans. However, the aforementioned datasets typically exhibit a bias towards standing poses due to the complicated capture procedure and cannot afford for large-scale data collection.  

\noindent \textbf{Multi-view Human Capturing Datasets.} Multi-view capture holds an indispensable role in computer vision, serving as a fundamental technique for AR/VR and 3D content production. Prior works~\cite{vlasic2008articulate, vlasic2009dynamic} present multi-view stereo systems to collect multi-view human images and apply multi-view constraints to reconstruct 3D virtual characters.  Human3.6M~\cite{ionescu2013human3} captures numerous 3D human poses using a marker-based motion capture system from 4 cameras. MPI-INF-3DHP~\cite{mehta2017monocular} annotates both 3D and 2D pose labels for human motion capture in a multi-camera studio.  CMU Panoptic~\cite{joo2015panoptic} presents a massively multiview system consisting of 31 HD Cameras to capture social interaction and provides 3D keypoints annotations of multiple people. HUMBI~\cite{yu2020humbi} collects local human behaviors such as gestures, facial expressions, and gaze movements from multiple cameras. AIST++~\cite{li2021ai,tsuchida2019aist}  is a dance database that contains various 3D dance motions reconstructed from real dancers with multi-view videos. These datasets primarily focus on human activity motions ranging from daily activities to professional performances, rather than factors related to identity,  cloth texture and body shape diversity. With the recent progress of neural rendering techniques, NHR~\cite{wu2020multi}, ZJU-Mocap~\cite{peng2021neural}, Neural Actor~\cite{liu2021neural, habermann2021real, habermann2020deepcap} and THuman4.0~\cite{zheng2022structured} present their multi-view human dataset for evaluating the proposed human rendering algorithms. HuMMan~\cite{cai2022humman} and Genebody~\cite{cheng2022generalizable} expand the diversity of pose actions and body shapes for human action recognition and modeling. ActorsHQ~\cite{isik2023humanrf} uses dense multi-view capturing for photo-realistic novel view synthesis but is limited to 16 motion sequences and 8 actors.  
Recently, with the presence of the large-scale synthetic data and real captures from Objaverse~\cite{deitke2023objaverse, deitke2023objaversexl} and MVImgNet~\cite{ yu2023mvimgnet}, several methods~\cite{liu2023zero, hong2023lrm} have made remarkable strides in the direction of open-world 3D reconstruction and generation. The concurrent work, DNA-Rendering~\cite{cheng2023dna} emphasizes the comprehensive benchmark functionality, but it encounters challenges in expanding the dataset to a larger scale due to the consideration of unusual human-object interactivity and clothes texture complexity. Differing from these efforts, we take a significant step forward in scaling up the human subjects and outfits, leading to the creation of \thename++, the multi-view human capture dataset on the largest scale.

% \input{background.tex}

% \section{2D Natural Images}
% \label{sec:2D}
% \input{2D-V2}
\vspace{-1mm}
\section{\thename++}
\label{sec:mvhuman}
\begin{figure}[t]
\begin{center}
\includegraphics[width=0.99\linewidth]{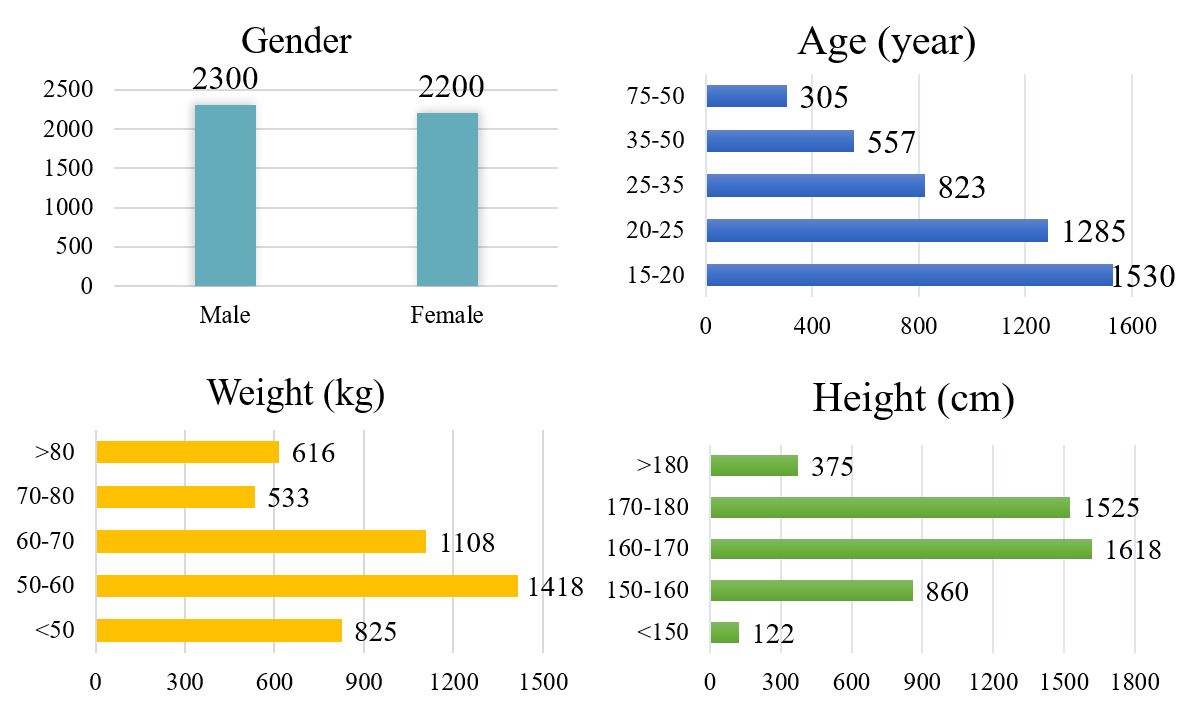}
\end{center}
\vspace{-1mm}   %added by chenghong
\caption{
\textbf{The distribution of performers' attributes.} The gender, age, weight, and height of performers are recorded and carefully controlled. The statistical analysis of these attributes reflects a diverse range among the performers involved in \thename++.}
\label{fig:0_human}
\vspace{-3mm}
\end{figure}

In this section, we provide a comprehensive overview of the core features of \thename++, with a focus on dataset construction. 
We discuss the hardware capture system, data collection arrangements, dataset statistics, and data pre-processing. 
Sec. \ref{data_sys} provides an illustration to the fundamental aspects of the data acquisition system. 
This part specifically outlines the key components of the hardware capture system and its capabilities. Sec. \ref{data_cap} delves into the actual data acquisition process, providing detailed information on personnel arrangement and the protocols followed during data collection. 
This section elucidates the steps taken to ensure the accuracy and consistency of the acquired data.
Finally, in Sec. \ref{data_ann}, we present a comprehensive framework that combines manual annotation and existing algorithms to obtain diverse and rich annotations for \thename++. This framework enhances the applicability of our dataset for various research purposes.

\begin{figure}%[htb]
\begin{center}
\includegraphics[width=0.99\linewidth]{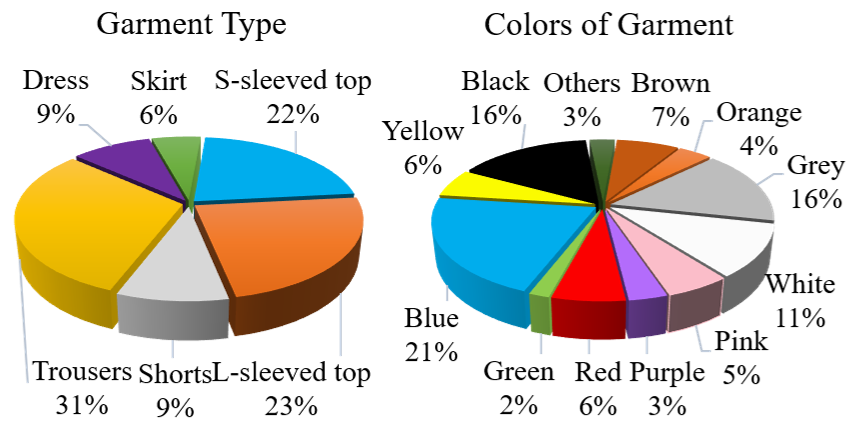}
\end{center}
\vspace{-3mm} %added by chenghong
\caption{
\textbf{The garment type and color distribution of outfits of performers.} Diverse colors and types of dressing are required for each invited performer. The statistical results show the wide coverage of daily clothes.
}
\vspace{-3mm}
\label{fig:0_garment}
\end{figure}

\subsection{Multi-view Synchronized Capture System}\label{data_sys}
We collected all the data using two sets of synchronized indoor video capture systems. The primary framework of the capture system consists of 48 high-definition industrial cameras with a resolution of 12MP. These cameras are arranged in a multi-layer structure resembling a 16-sided prism, as shown in Fig. \ref{fig:teaser}. 
The collection system has approximate dimensions of 2.4 meters in height and a diameter of roughly 4.5 meters.
Each prism within the system is equipped with three 4K high-definition industrial cameras positioned at different heights.
The lenses of each camera are meticulously aligned towards the center of the prism. 
%\lch{This arrangement allows us to capture the subject's body movements and intricate texture details of their clothing from 48 distinct viewing angles.
To ensure clear image capture from different perspectives, we have placed light sources at the center of each edge of the system.
%These light sources facilitate optimal lighting conditions during data collection.}
During the data collection process, the frame rate of all cameras is set to 25 frames per second, enabling the capture of smooth and detailed motion sequences.

% The second capture system consists of 24 high-definition industrial cameras which are evenly distributed on 16 pillars in a two-layer structure. The collection system has approximate dimensions of 2.2 meters in height and roughly 4.3 meters in diameter.
% The lenses of each camera are meticulously aligned towards the center of the prism. 
% To ensure clear image capture from different perspectives, we place light sources at the center of each edge of the system.
% During the data collection process, the frame rate of all cameras is set to 30 frames per second, enabling the capture of smooth and detailed motion sequences.

The secondary system consists of 24 high-definition industrial cameras with a resolution of 5MP, evenly distributed across 16 pillars in a two-layer structure. This system measures approximately 2.2 meters in height and 4.3 meters in diameter. Similar to the primary system, the lenses are aligned toward the center, and light sources are placed at each edge to ensure optimal lighting. The cameras in this system operate at 30 frames per second, further enhancing the quality of motion sequence capture.

% \begin{figure}[tb]
% \begin{center}
% \includegraphics[width=0.85\linewidth]{figures/24views_mvs.pdf}
% \end{center}
% \vspace{-3mm}
% \caption{
% \textbf{The visualization of the second multiview synchronized capture system.} Our second capture system consists of 24 industrial cameras with a resolution of 2448$\times$2048. 
% }
% \label{fig:24views_mvs}
% \vspace{-3mm}
% \end{figure}

\begin{figure}[tb]
\begin{center}
\includegraphics[width=1.0\linewidth]{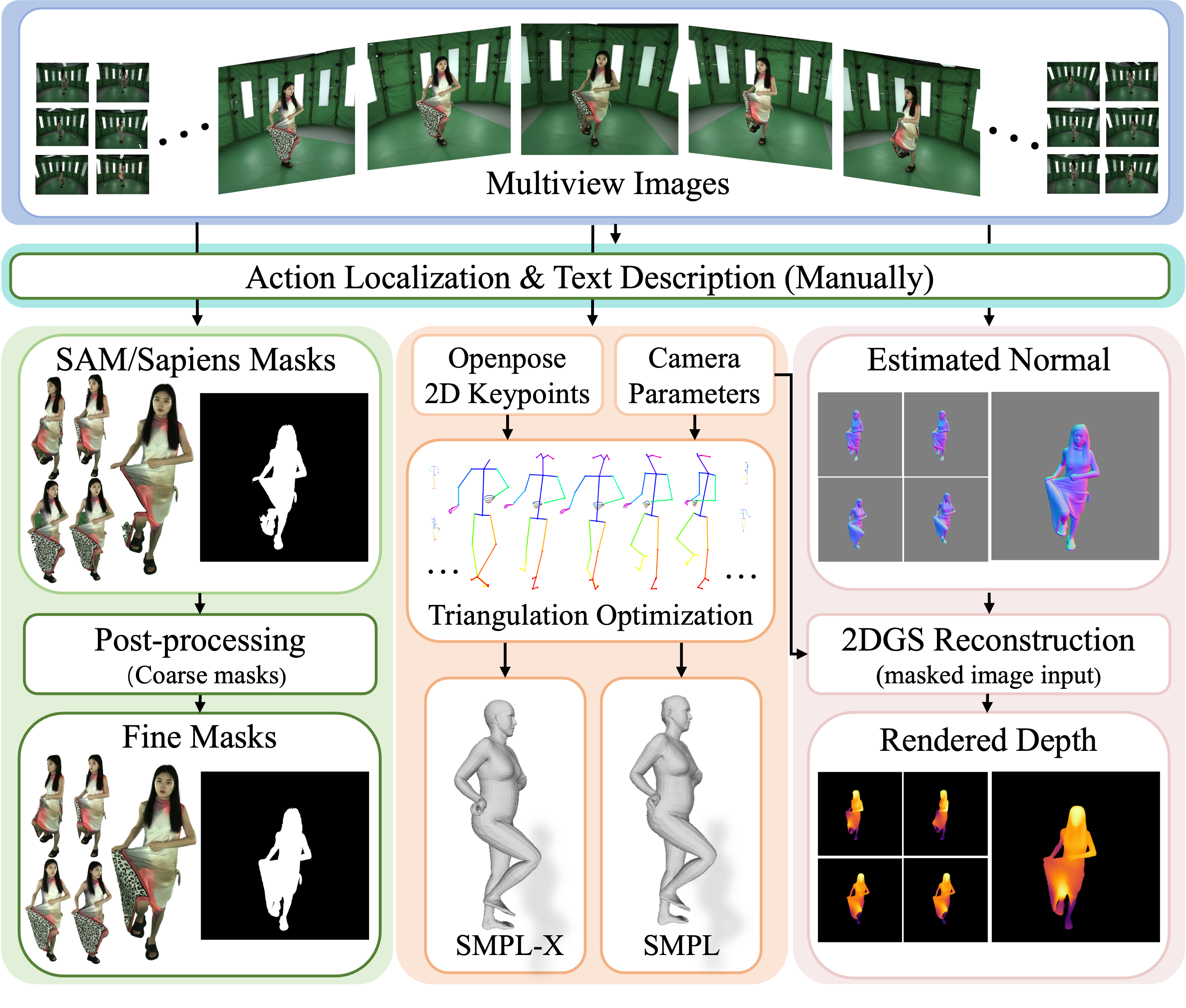}
\end{center}
\vspace{-3mm}
\caption{
\textbf{Data annotation pipeline.} The manual and automatic annotation pipeline for action localization, text description, masks, 2D/3D keypoints, parametric models, normal maps and depth maps.}
\label{fig:process}
\end{figure}

\subsection{Data Capture and Statistics}\label{data_cap}

\noindent\textbf{Data Capture} 
To capture the wide range of dressing habits observed in people's daily lives, we establish a comprehensive process for performer recruitment and data collection.
Specifically, at regular intervals, we release targeted recruitment requests to the public based on the statistics derived from the already collected clothing data. 
This strategy aims to enhance the diversity of clothing styles and colors for more reasonable human data distributions to achieve more reasonable human data distributions.
% \lch{In accordance with the clothing requirements, each performer is assigned a specific outfit to wear and is instructed to bring a extra set of clothing to the capture system.}  
In accordance with the clothing requirements, each performer is instructed to bring two sets of clothing to the capture system.  Prior to the beginning of the capturing, performers randomly select 12 sets of actions from a predefined pool of 500 actions. 
Subsequently, they enter the capture system and sequentially perform the first six sets of actions, following instructions provided by the collection personnel. Each action is performed at least once on both the left and right sides for complete execution of the human performance capture.
Upon completing the sixth set of actions, the performer finishes the first collection session by extending their hands to an A-pose and rotating in place twice. Subsequently, the performer changes outfit and repeats the same process to complete the remaining six sets of actions with rotations in place.

\noindent\textbf{Data Statistics}
The essential statistics of our dataset are shown in Fig. \ref{fig:0_human} and Fig. \ref{fig:0_garment}.
%In order to ensure a comprehensive representation of various attributes of everyday clothing in our dataset, we meticulously designed the collection process, paying close attention to garment styles, colors, as well as the age and gender of the performers.
%Furthermore, we made concerted efforts to maximize the diversity of performed actions, resulting in the inclusion of 500 distinct actions in our dataset.
\thename++~comprises a total of 4,500 unique identities with a equitable distribution of 2,300 male and 2,200 female individuals, ensuring a balanced representation of genders.
%and 9,000 sets of clothing styles. 这里再写服装感觉不太顺 
%Recognizing that age can influence the quality of performance actions, we implemented specific age restrictions during the recruitment process. 
Participants are required to fall within the age range of 15 to 75 years old. 
This age range is chosen to encompass a wide spectrum of performers while considering the potential impact of age on the quality and capabilities of their actions.
Conversely, no restrictions are imposed on performers' weight or height, as these variables are deemed to have minimal impact on the data collection process. 
By not imposing such limitations, we aim to capture a more diverse and realistic representation of subjects in the dataset, allowing for a broader range of body types and proportions.
Our dataset boasts the largest number of unique identities and garment items when compared to existing multi-view human dataset . 
It encompasses a wide range of everyday clothing styles and colors that are commonly available in real-world scenarios.

%Detailed statistics of actors are shown in ~\cref{fig:0_human}.

%Due to the inherent variations in garment styles and colors between males and females in real life, such as men's preference for black, white, and gray colors, and the tendency for women's garments to be more extravagant and diverse in style, achieving diversity and balance in the dataset presented significant challenges. 
%To effectively enhance the diversity of clothing styles and colors in the dataset, we implemented regular modifications to the clothing requirements throughout the recruitment process. 
%Performers were strictly instructed to bring appropriate %clothing that adhered to the specified requirements.
%The final statistical results are depicted in the figure, with clothing roughly categorized into six major classes and colors classified into twelve major classes. 
%It is important to note that slight differences in color classification may exist due to individual variations in color perception.

\begin{figure}[tb]
\begin{center}
\includegraphics[width=0.9\linewidth]{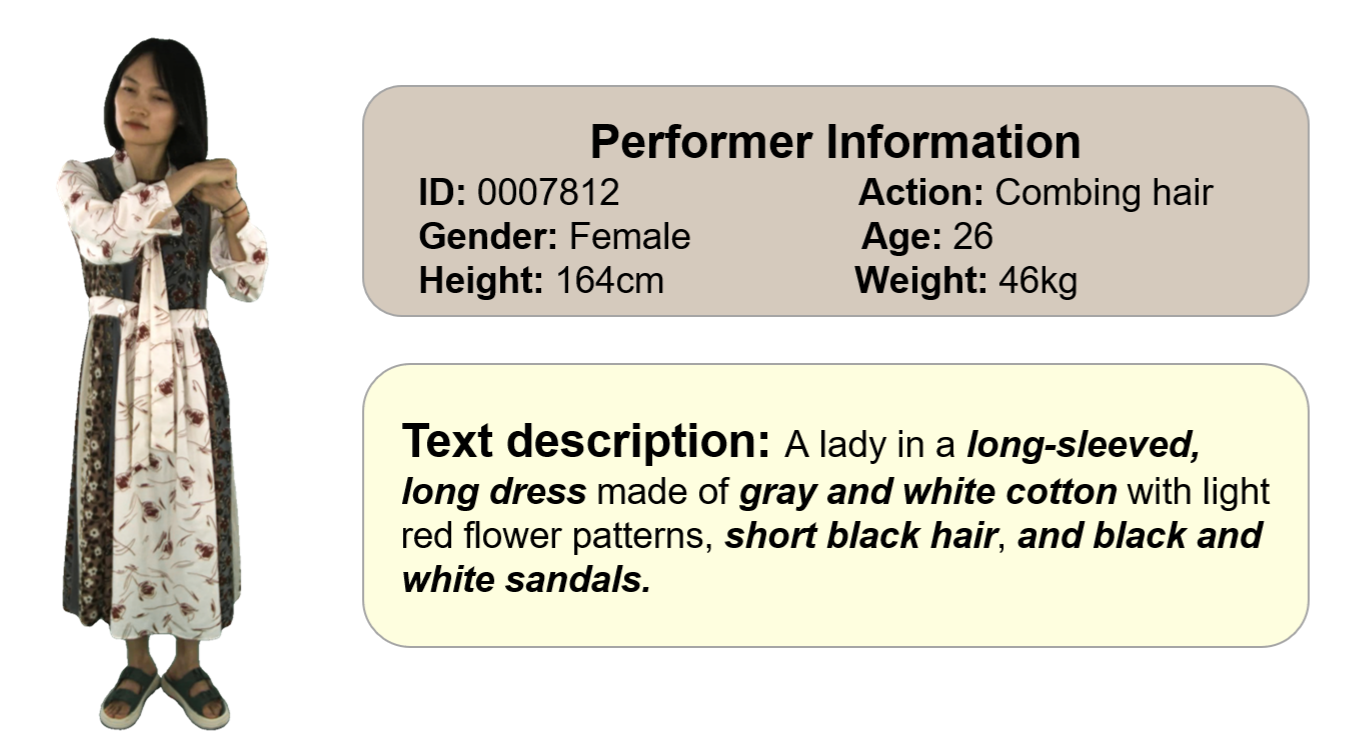}
\end{center}
\vspace{-3mm}
\caption{
\textbf{A text description example.} The description contains various information, such as age, height, garment and hairstyle. }
\label{fig:text_desc_small}
\vspace{-3mm}
\end{figure}

\subsection{Data Annotation}\label{data_ann}
To enable the advancement of applications in 2D/3D human understanding, reconstruction and generation, our dataset offers comprehensive and diverse annotations alongside the raw data.
These annotations include action localization, attribute description, human masks, camera calibrations, 2D/3D skeleton, and parametric model fitting. Tab. \ref{tab:GPU_hours} shows GPU hours for data processing. The annotation pipeline, as depicted in Fig. \ref{fig:process}, provides an overview of the entire process.

\begin{table}[htbp]
    % \resizebox{\columnwidth}{!}{
    \centering
    \setlength{\tabcolsep}{8pt} % 增加列间距
    {
    
    \begin{tabular}{c|cccc}
        \toprule
        Annotation & Mask  & SMPLX (SMPL) & Normal & Depth \\
        \midrule
        GPU hours & 1954.6 & 2456.8 & 1345.0 & 8469.2 \\

        % Ours &\cellcolor{deeppeach}27.568 & \cellcolor{deeppeach}0.954 & \cellcolor{lightgray} 0.0682 \\
        \bottomrule
    \end{tabular}
    }
    \caption{
    % Performance of different models. 
    GPU hours for data processing
    }
    \label{tab:GPU_hours}
\end{table}

\noindent\textbf{Manually Annotation} 
Before capturing human data, we collect the cloth color and dress type of each performer in the registration table for manual textual description. We carefully record the essential details of each identification encompassing crucial information such as gender and age.  Furthermore, we employ manual labeling to furnish text descriptions of the performers’ hairstyles and shoes, as well as each outfit, including clothing color, style and material. Fig.~\ref{fig:text_desc_small} provides a visual representation.
During the data collection process, we ensure a continuous flow as performers execute a sequence of six distinct actions along with in-place rotations. 
Subsequently, after the recording session, we manually mark the breakpoints for each action, accurately documenting the start and end  of each action sequence.  

\noindent\textbf{Camera Calibration}
We utilized a commercial solution based on CharuCo boards to achieve fast and efficient camera calibration. 
Specifically, we position a CharuCo patterned calibration board at the central location of the capture studio.
This ensures that each camera can capture a clear and complete view of the calibration board. With the aid of specific software, we obtain the intrinsic, extrinsic parameters and distortion coefficient for each camera. 
%Moreover, recognizing the potential for performers to inadvertently come into contact with the capture studio or cameras during their entry or execution of actions, we implement a calibration process at the beginning, middle, and end of each day. 
%This procedure aim to account for any potential changes in camera parameters. 
%To evaluate the necessity of data recapturing, we compute the L1 distance between the camera parameters.  
Moreover, recognizing the potential for performers to inadvertently come into contact with the capture studio or cameras during their entry or execution of actions, we implement a calibration process at the beginning, middle, and end of each day. 
This procedure aims to account for any potential changes in camera parameters. 
We also carefully adjust other parameters, such as lighting, exposure, and camera white balance to capture high-quality data.

\begin{figure}[tb]
\begin{center}
\includegraphics[width=1.0\linewidth]{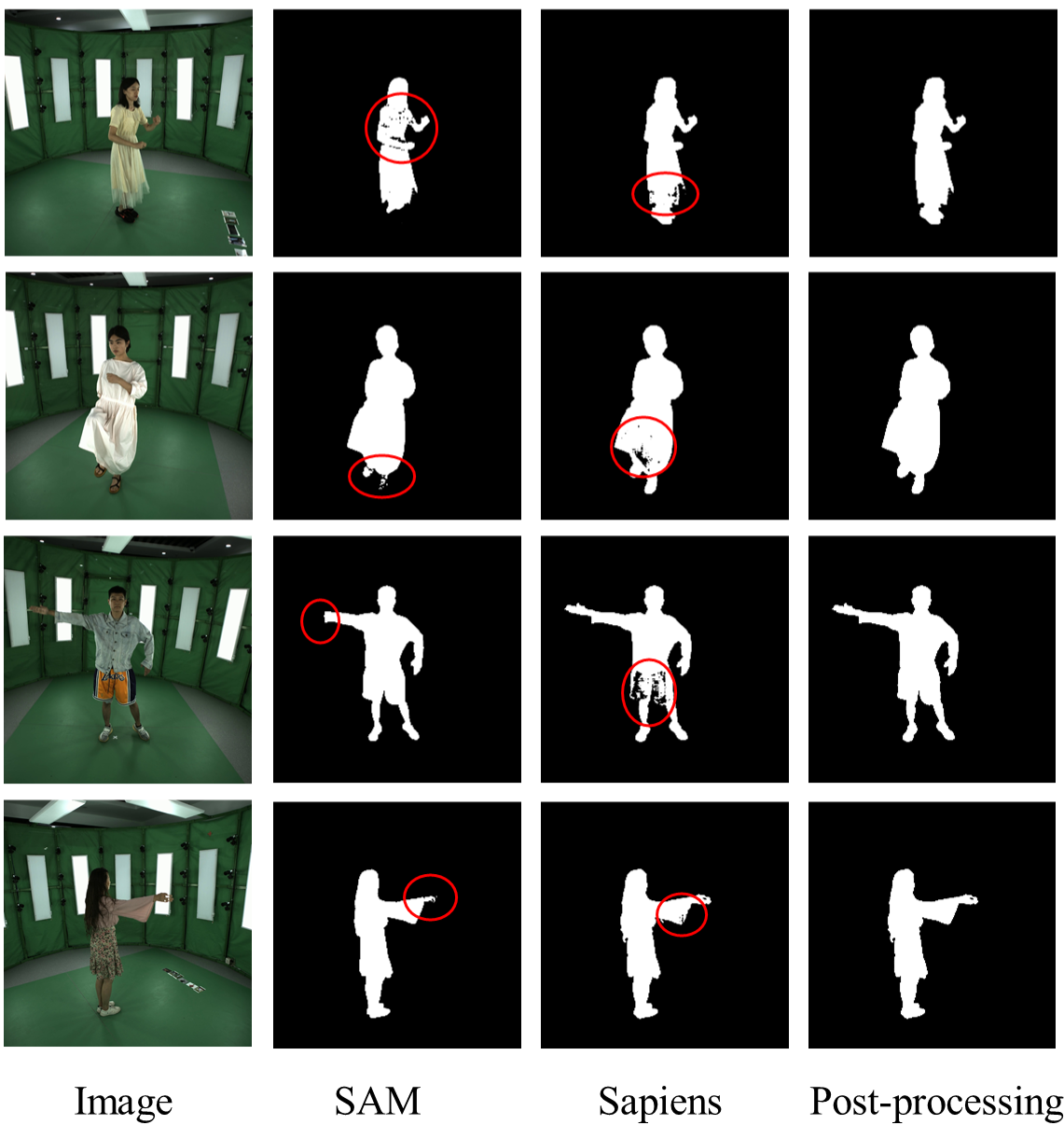}
\end{center}
\vspace{-2mm}
\caption{
\textbf{Mask processing visualization.} From left to right in each column are the input image, SAM segmentation result, Sapiens segmentation result, and final mask after post-processing.}
\label{fig:mask}
\vspace{-2mm}
\end{figure}

\noindent\textbf{Human Mask Segmentation}
\thename++~comprises approximately 645 million images of individuals captured from various perspectives. 
Manual segmentation of such a massive image collection is obviously infeasible. In our conference paper~\cite{xiong2024mvhumannet}, we propose a hierarchical automated image segmentation approach based on off-the-shelf segmentation algorithms. Nonetheless, SAM cannot generalize very well for human body segmentation. With the recent introduction of the Human Foundation Model, Sapiens~\cite{khirodkar2024sapiens}, we leverage its powerful segmentation capabilities to generate masks for images where the human segmentation accuracy of SAM is insufficient. We observe that Sapiens performs well for tight-fitting clothing, accurately capturing hand contours. However, for loose-fitting clothing, the masks generated by Sapiens often exhibit noticeable artifacts. To address this issue, we propose a post-processing method to enhance mask quality as illustrated in \cref{alg:mask}. Note that under this paradigm, we significantly reduced the quantity of masks needed to be manually inspected and labeled. The mask visualization results are shown in Fig. \ref{fig:mask}.

% To tackle this challenge, we propose a hierarchical automated image segmentation approach based on off-the-shelf segmentation algorithms. 
% Our approach follows a coarse-to-fine segmentation strategy. 
% Initially, we employ the RVM~\cite{rvm} to obtain efficient rough segmentation results. Subsequently, the rough segmentation outcomes are utilized to generate a bounding box of the performer, which serves as a prompt for the SAM~\cite{Kirillov_SAM}  to produce higher-quality masks. 
% In~\cref{fig:process}, the bottom-left region presents a comparison between the coarse and fine segmentation results. 
% Notably, the masks generated by SAM exhibit significant superiority to those generated by RVM.
% ~\cref{fig:mask} shows the xxxxxxx. 

% Nevertheless, RVM remains crucial as it provides a rough bounding box, ensuring that the fine stage SAM segmentation primarily focuses on the individual rather than other elements.  

\begin{algorithm}[b]
    % \footnotesize
    \small
    \caption{Procedure of Mask Enhancement}
    \label{alg:mask}
    \renewcommand{\algorithmicrequire}{\textbf{Input:}}
    \renewcommand{\algorithmicensure}{\textbf{Output:}}
    
    \begin{algorithmic}[1]
        \REQUIRE $M$ (masks from SAM or Sapiens output), $T_{hold}$ (Threshold of max hole area needed to be filled)  %% 输入
        \ENSURE $M_{out}$ (final enhanced mask)  %% 输出
        
        \STATE Extract outer contours $C = \{C_1, C_2, ..., C_n\}$ from $M$
        \STATE Compute contour sizes $S = \{s_1, s_2, ..., s_n\}$
        \STATE Identify the largest outer contour $C_{max}$ with size $s_{max}$ and discard other contours as $M$        
        \STATE Extract inner contours as holes $H = \{H_1, H_2, ..., H_m\}$ inside $C_{max}$
        \IF {No holes}
            \STATE Return mask $M$ as $M_{out}$
        \ELSE
            \IF{size of all $H_i < T_{hole}$}
                \STATE Fill all $H_i$ in $M$ and return $M$ as $M_{out}$
            \ELSE
                \STATE Fill those $H_i < T_{hole}$ in $M$ and perform manual inspection
                \IF{obvious missing regions detected}
                    \STATE Perform union of SAM and Sapiens $M$
                    \IF{Still obvious error regions detected}
                        \STATE Return manually labeled mask as $M_{out}$
                    \ENDIF
                \ENDIF
            \ENDIF
        \ENDIF
                
        \STATE Output enhanced mask $M_{out}$
    \end{algorithmic}
\end{algorithm}

\begin{figure*}
\begin{center}
\includegraphics[width=1.0\textwidth]{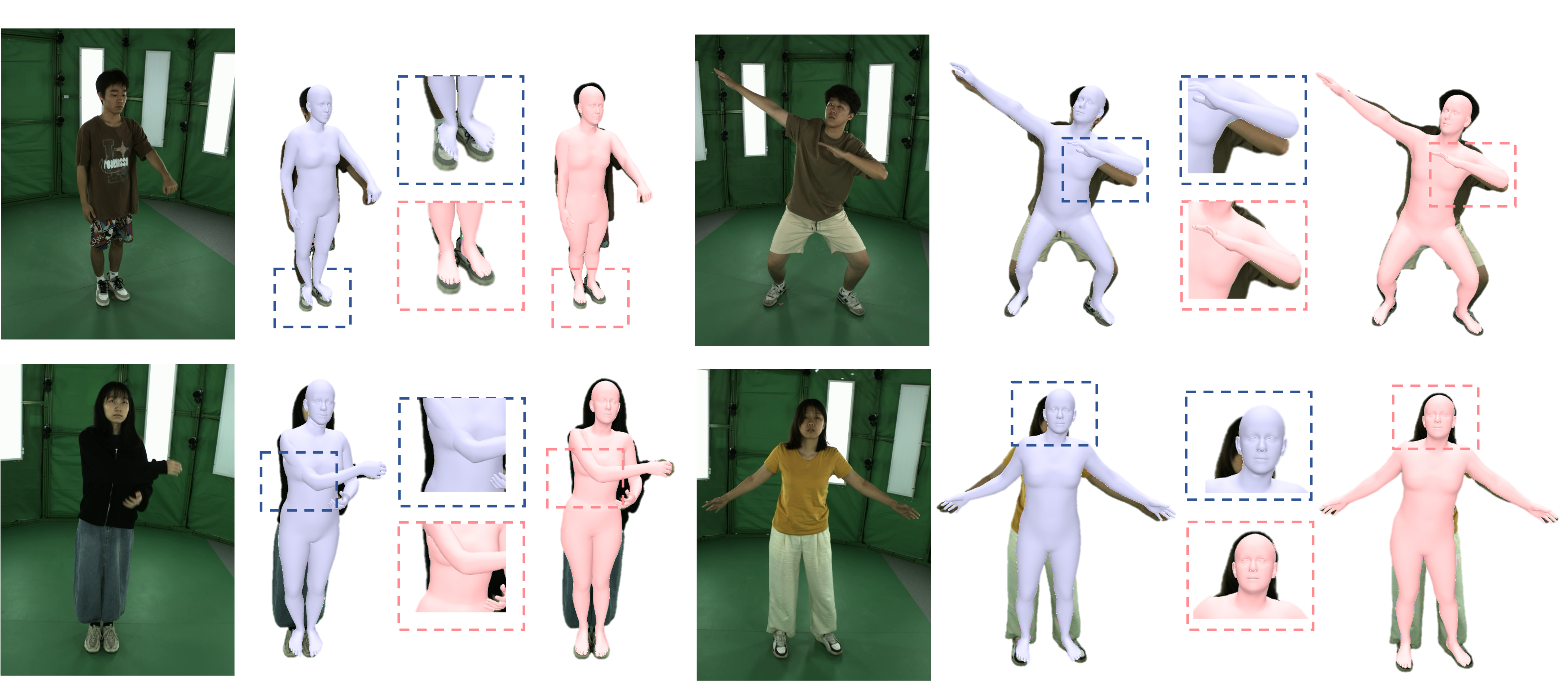}
\end{center}
\vspace{-3mm}
\caption{
\textbf{SMPLX comparsion results.} The zoom-in boxes with blue dot lines show the annotation quality before optimization and the pink ones show quality improvements. Previous SMPLX estimation results show ankle twisting and self-intersection artifacts in the left column images, as well as misalignment in the right column images. In contrast, our optimized pipeline incorporates a body pose prior to regularize human pose estimation, effectively addressing these limitations.}
\vspace{-2mm}

\label{fig:smplx}

\end{figure*}

\noindent\textbf{2D/3D Skeleton and Parametric Models}
Following the previous works~\cite{cai2022humman, cheng2023dna, cheng2022generalizable} and with the goal of facilitating extensive research and applications in 3D digital human community, we conducted pre-processing on the entire dataset to obtain corresponding 2D/3D skeletons and two parameterized models. 
The processing pipeline is visually depicted in the middle-bottom part of Fig. \ref{fig:process}. 
Specifically, we employed the OpenPose~\cite{cao2017openpose} to predict 2D skeletons for each frame of the images. By leveraging the calibrated camera parameters and multi-view 2D skeletons, we employ the multi-view triangulation algorithm to derive 3D keypoints. In our conference paper~\cite{xiong2024mvhumannet},  due to the large-scale multi-view image collection,  we use the open-source toolbox EasyMocap~\cite{easymocap}, which provides efficient runtime capabilities, to optimize SMPL/SMPLX parameters with the constrains of multi-view 2D keypoints and 3D skeletons. However, we find that EasyMocap only registers the SMPL/SMPLX using 3D keypoints without body pose prior, which can easily lead to unrealistic joint distortions, such as elbows bending backward and ankle twisting. Thus we incorporate the body pose prior to refine the human pose in the latent space of VPoser~\cite{pavlakos2019expressive}, a variational human pose prior trained on the AMASS dataset~\cite{AMASS:ICCV:2019}.  The visualization results of SMPLX comparison are shown in Fig. \ref{fig:smplx}.

\begin{figure}[htbp]
\begin{center}
\includegraphics[width=0.95\linewidth]{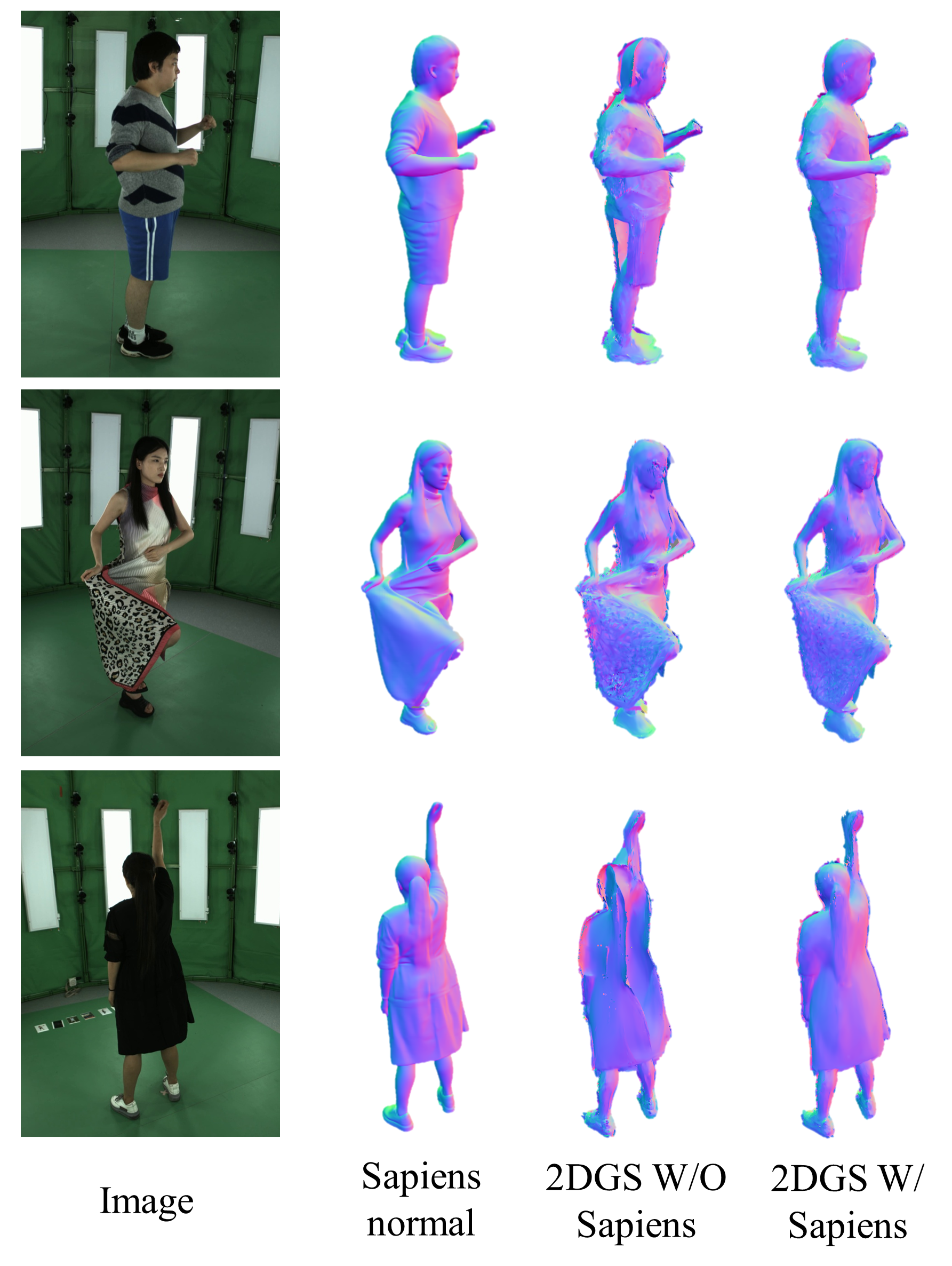}
\end{center}
\vspace{-1mm}
\caption{
\textbf{Normal visualization.} From left to right of each column are the original image, Sapiens estimated normal, 2DGS rendered normal without Sapiens normal, and 2DGS rendered normal with Sapiens normal input.}
\label{fig:normal}
\end{figure}

\noindent\textbf{Normal Maps}
Normal maps are crucial for high-fidelity 3D human reconstruction as they enhance the representation of surface details, such as garment wrinkles, and further improve the overall quality of reconstructed models~\cite{xiu2022icon, saito2020pifuhd}. Moreover, normal information facilitates the integration of photometric cues from multi-view images by compensating for missing details in low-texture or highly illuminated regions, thus improving pixel intensity matching across views in multi-view reconstruction~\cite{long2024wonder3d, ye2024stablenormal}. However, ground-truth normal map is unavailable for real-capture mutli-view human data, we attempt to use the 2D human normal foundation model Sapiens~\cite{khirodkar2024sapiens} to generate pseudo labels for normal maps. We leverage our generated normal maps to regularize the human surface reconstruction method following 2DGS~\cite{huang20242d}. The visualization of
normal rendering results are shown in Fig.~\ref{fig:normal}.

\begin{figure}[htbp]
\begin{center}
\includegraphics[width=0.99\linewidth]{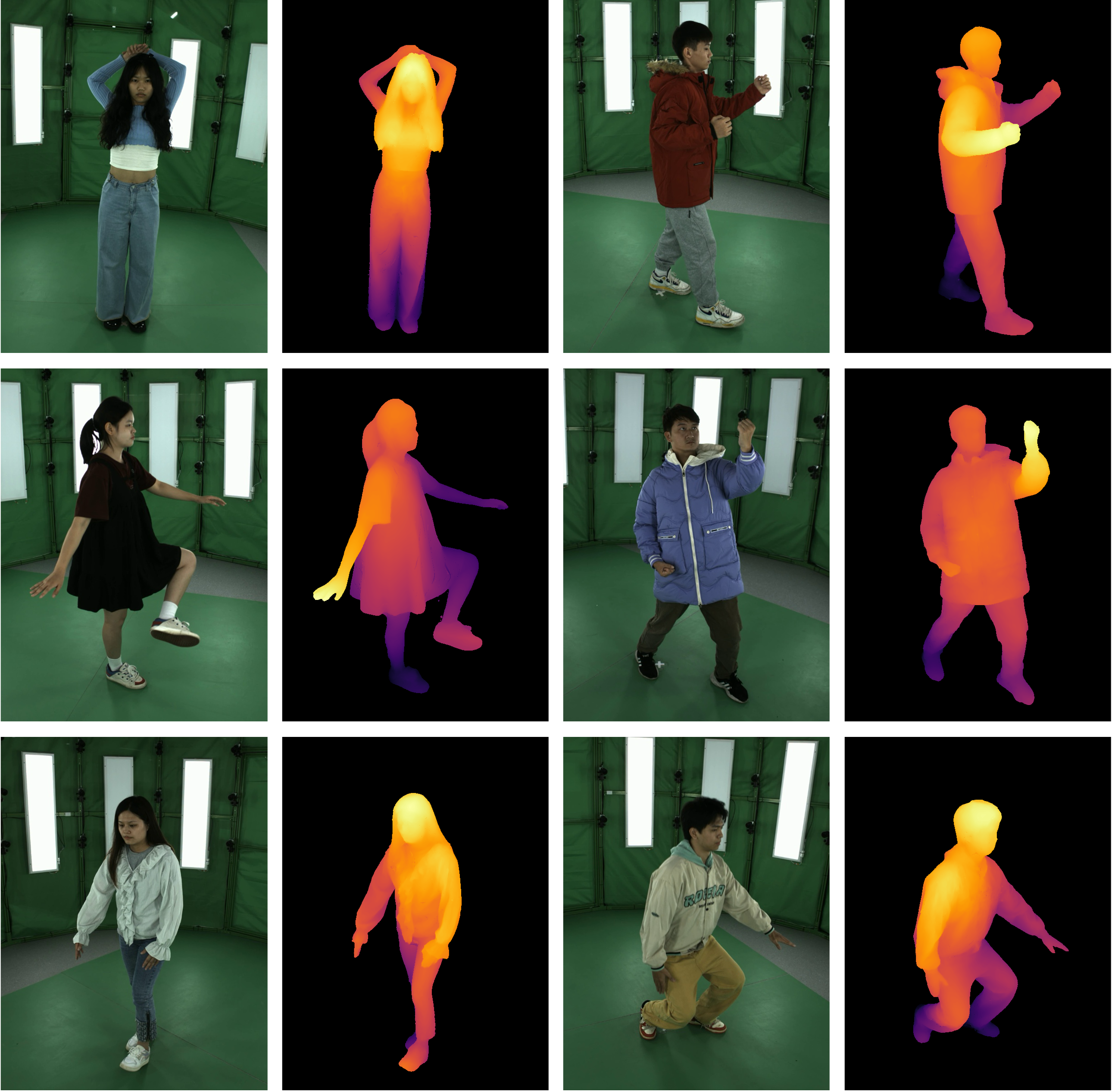}
\end{center}
\vspace{-2mm}
\caption{\textbf{Depth visualization.}
Visual results of depth maps rendered from normal-refined 2DGS.}
% From left to right of each row are the original image, SAM result, Sapiens result and final mask after post-processing.
\label{fig:depth}
\end{figure}

\noindent\textbf{Depth Maps}
Depth maps are also essential data types in human capture datasets,  as they directly record the 3D structural information of the human body. Unlike RGB images, depth maps are not affected by factors such as lighting or texture, making them a more reliable source of geometric information. For 3D human reconstruction, depth map provides reliable geometric input for neural networks to infer accurate 3D shapes~\cite{chibane2020implicit}. Furthermore, depth information can capture fine details such as cloth wrinkles, which can serve as a supervisory signal to constrain the 3D human geometry, and further improve the quality of reconstruction~\cite{zheng2024gps, hu2024eva}. However, our multi-view human capture system is only equipped with calibrated RGB cameras, which cannot directly obtain depth maps. Inspired by the aforementioned normal-refined 2DGS results, we use 2D Gaussian primitives and multi-view camera parameters to render human depth maps for each view. The visualization results of our processed depth map are show in Fig.~\ref{fig:depth}. Additionally, we conduct several experiments to demonstrate that the depth maps generated from 2D Gaussian primitives can serve as pseudo-label supervision for human gaussian rendering and unconstraint reconstruction.

% \section{Video Understanding}
% \label{sec:video}
% \input{video_V5.tex}

% \section{Vision + Language}
% \label{sec:VL}
% \input{VL_V5}

% \section{3D Data Analysis}
% \input{representation-V2}
% \input{understanding-V2}
% \input{generation-V2}

% \vspace{-3mm}
% \section{Medical Image Analysis}
% \label{sec:medical}
% \input{medical.tex}

% Please refer to the supplementary material for the details of this section, including brain activity investigation, disease diagnosis, and anatomy segmentation. 

\section{Experiments}
In this section, we present a comprehensive series of exploratory experiments conducted in the human action understanding, reconstruction, and generation tasks. Specifically, Sec. \ref{action_task} highlights experiments focused on view-consistent action recognition. As the dataset expands from single-view 2D data to multi-view 3D data, existing algorithms may encounter new challenges. 
In Sec. \ref{nerf_task} and Sec. \ref{general_gs}, we demonstrate experiments on generalizable NeRF and gaussian splatting reconstruction approaches, highlighting the augmented model performance and generalization capabilities resulting from the increased availability of data. Sec. \ref{per_subject_gs} emphasizes the rendering quality comparisons of Animatable Gaussians between using the original and new SMPLX parameters. 
At last, in Sec. \ref{text_task}, Sec. \ref{avatar_task} and Sec. \ref{dust3r_task}, we delve into recent research tasks, specifically text-driven view-unconstrained image generation,  3D human avatar generative model, multi-view human images generation and reconstruction from unconstraint human images.
Taking into account the size of the dataset, hardware limitations, and data annotation constraints, we performed experiments utilizing 62{\%} of the available data. More precisely, we employed 2800 identities, each representing a unique set of attire, amounting to a total of 5500 sets. Within this subset, 10{\%} of the data was reserved exclusively for testing purposes.

\subsection{View-consistent Action Recognition }\label{action_task}
\thename++~provides action labels with 2D/3D skeleton annotations, which can verify its usefulness on action recognition tasks. 
To simulate real-world scenarios, we employed single-view 2D skeletons as input and conducted tests on a multi-view test set that accurately represented real scenes. 
Our experimentation involved 8 viewpoints spaced at 45-degree intervals. 
The training data encompassed approximately 4000 outfits, while the testing data included 400 outfits, covering a total of 500 action labels. 
The results, presented in Tab. \ref{tab:action_experiments}, reveal that the accuracy of action estimation was notably low for a single viewpoint, achieving a top-1 accuracy of only around 30\%. 
However, as the number of input viewpoints increased, the accuracy of action estimation exhibited a significant improvement, peaking at 78.19\%. 
Given that the dataset covers a comprehensive range of daily full-body actions, we possess confidence in its efficacy for facilitating diverse understanding tasks.
Considering the challenges associated with acquiring 3D skeletons in everyday life, see supplementary for the results of 3D skeleton-based action recognition.

%Specifically, we train state-off-the-art graph-based methods~\cite{zhou2023FRHEAD, chi2022infogcn, chen2021CRTGCN}on our dataset. 
%The view-consistent 2D action recognition results illustrated in the Tab~\ref{tab:action}. 
%We selected one camera viewpoint at intervals of 45 degrees, resulting in a total of 8 distinct viewpoints. 
%Then, we progressively augmented the number of viewpoints to
%train the network training, and the test set comprised 8 cameras. 
%Notably, the results exhibited a pronounced enhancement in action prediction accuracy as the number of training data viewpoints increased, aligning precisely with our initial expectations.

\begin{table}%[t]
\centering
\small
%\vspace{-2mm}
\def\arraystretch{1} \tabcolsep=0.4em 
\begin{tabular}{r|r|c|c|c}
    \toprule
    & \makecell{Train\\ views}& CTR-GCN\cite{chen2021CRTGCN} & InfoGCN\cite{chi2022infogcn} & FR-Head\cite{zhou2023FRHEAD}  \\
    \midrule
    \makecell{Top-1\\(\%)$\uparrow$ }& 
    \makecell{1-view\\ 2-views\\  4-views\\ 8-views} &  
    \makecell{33.85\\60.33\\72.16\\ 76.73 }  & 
    \makecell{25.23\\55.89\\73.59\\ 76.55}  &  
    \makecell{30.25\\59.16\\71.74\\ 78.19}  \\
    \midrule
    \makecell{Top-5\\(\%)$\uparrow$ }& \makecell{1-view\\ 2-views\\ 4-views\\ 8-views} & 
    \makecell{51.08\\80.09\\88.32\\ 91.34}  & 
    \makecell{37.14\\75.00\\89.02\\ 91.00}  &  
    \makecell{50.59\\78.80\\88.67\\ 92.45}  \\
    \bottomrule
\end{tabular}
\vspace{-1mm}
\caption{Performance comparison of skeleton-based action recognition SOTA methods on \thename++. With the increase of the views, the accuracy of the action prediction increases together.}
    \label{tab:action_experiments}
\end{table}

\subsection{NeRF Reconstruction for Human}\label{nerf_task}
\begin{figure}[t]
\begin{center}
\includegraphics[width=0.99\linewidth]{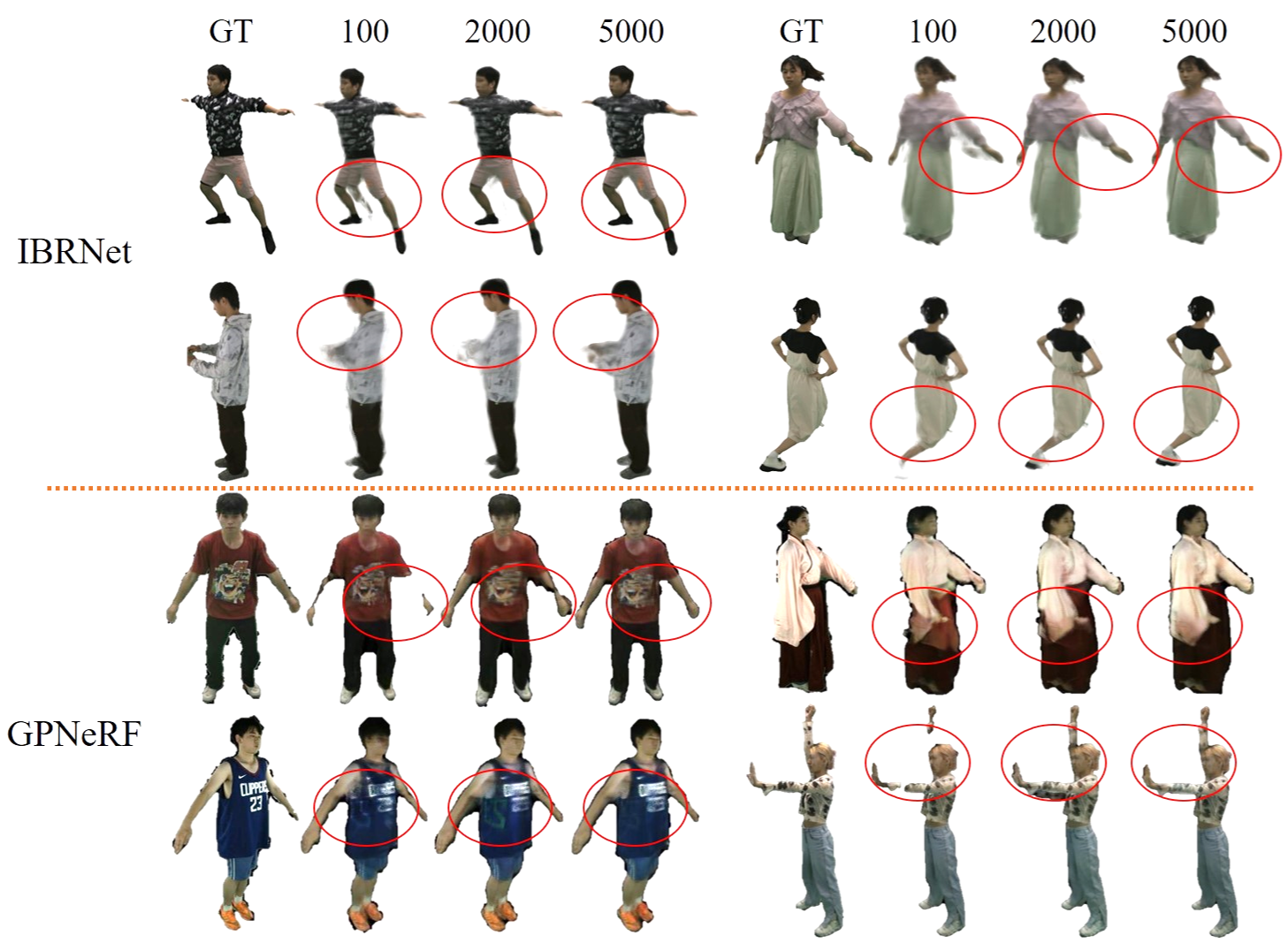}
\end{center}
\vspace{-3mm}
\caption{\textbf{The novel view synthesis results of IBRNet and GPNeRF on unseen data of \thename++.} \textbf{GT} means ground truth. The number of \textbf{100}, \textbf{2000}, and \textbf{5000} indicate the respective quantities of outfits utilized during the training process.}

\label{fig:nerf_vis}
\end{figure}

\thename++ can also be applied to NeRF reconstruction for human. Currently, human-centric methods, e.g. GPNeRF \cite{chen2022geometry}, are developed in the context of lacking multi-view human data and their performance is still far from satisfactory on more diverse testing cases. We hope our proposed \thename++ can motivate more extensive studies of generalizable NeRF for human with sufficiently large scale of data.  We empirically explore the performance of two distinct generalizable NeRFs methods,  IBRNet \cite{wang2021ibrnet} which is designed for general scenes and GPNeRF \cite{chen2022geometry} which relies on human prior (i.e. SMPL~\cite{loper2023smpl}), using varying amounts of data for training. In our experiment, both approaches utilize four evenly distributed views as input and inference the novel view results. The quantitative comparisons of the outcomes are presented in Tab.~\ref{tab:gene_human_nerf}, while the visualization results can be found in Fig.~\ref{fig:nerf_vis}. Experimental results confirm that as the training data increases, the model exhibits enhanced generalization capabilities for new cases, especially when facing rare poses and complex garments. Moreover, we provided empirical evidence that \thename++ can also serve for pretraining strong models, facilitating methods to perform better on out-of-domain scenarios. The corresponding results are presented in Tab.~\ref{tab:cross_dataset_nerf}. Please note that the quantitative results of IBRNet \cite{wang2021ibrnet} and GPNeRF \cite{chen2022geometry} cannot be directly compared, as they have different evaluation settings.

\begin{table}[tb]
    \resizebox{\columnwidth}{!}{
    % \large
    \begin{tabular}{c|ccc|ccc}
        \toprule
         %&  &  \multicolumn{3}{c}{\makecell{Static Scenes \\ w/ Ground Truth}} & \multicolumn{3}{c}{\makecell{Dynamic Scenes \\ w/o Ground Truth}} \\
        {\multirow{2}*{\begin{tabular}[c]{@{}c@{}}Number of \\ outfits\end{tabular}}}   &  \multicolumn{3}{c|}{IBRNet~\cite{wang2021ibrnet} }   & \multicolumn{3}{c}{GPNeRF~\cite{chen2022geometry}}   \\
         &  PSNR $\uparrow$ &  SSIM $\uparrow$ &  LPIPS $\downarrow$ &  PSNR  $\uparrow$ &  SSIM $\uparrow$ &  LPIPS $\downarrow$ \\
        \midrule
             100     & 26.05  & 0.9571  & 0.0555  &  23.27 & 0.8688 &  0.2077 \\
            2000    & 27.45  & 0.9638  & 0.0486  & 24.14 & 0.8779 & 0.2137 \\
            5000    & 29.00 & 0.9706 & 0.0377  & 24.69 & 0.8878 & 0.1961 \\
        \bottomrule
    \end{tabular}
    }
    \vspace{-1mm}
    \caption{\small \textbf{Quantitative comparison of generalizable NeRFs with different scales of data for training.} We compare the results of methods with human prior and without human prior. We refer human prior to the commonly used SMPL model.}
     % \vspace{-4mm}
    \label{tab:gene_human_nerf}
\end{table}

\begin{table}[tb]
    % \resizebox{\textwidth}{!}{
    \resizebox{\columnwidth}{!}{
    \begin{tabular}{c|ccc|ccc}
        \toprule
         %&  &  \multicolumn{3}{c}{\makecell{Static Scenes \\ w/ Ground Truth}} & \multicolumn{3}{c}{\makecell{Dynamic Scenes \\ w/o Ground Truth}} \\
          {\multirow{2}*{Method}}  &  \multicolumn{3}{c|}{IBRNet~\cite{wang2021ibrnet}} & \multicolumn{3}{c}{GPNeRF~\cite{chen2022geometry}} \\
          &  PSNR $\uparrow$ &  SSIM $\uparrow$ & LPIPS $\downarrow$ &  PSNR $\uparrow$  &  SSIM $\uparrow$ & LPIPS $\downarrow$ \\
        \midrule
          Train from scratch & 28.06 &  0.9679 & 0.0437        & 20.95 & 0.9049 & 0.1809\\
          w/o fintune & 27.48 & 0.9663 & 0.0440 & 20.15 & 0.8921 & 0.2050 \\
          w/ fintune  & 29.46 & 0.9734 & 0.0323 & 21.89 & 0.9252 & 0.1364  \\
        \bottomrule
    \end{tabular}
    }
    \vspace{-1mm}
    \caption{\small \textbf{Using \thename++ to pretrain a strong model}. We first train the representative methods on \thename++, and then finetune the trained models on the train set of HuMMan~\cite{cai2022humman}. We compare the performance of the finetuned models and models trained from scratch on the test set of HuMMan.}
    % \vspace{-0.4cm}
    % \vspace{-4mm}
        \label{tab:cross_dataset_nerf}
\end{table}
%\vspace{-5mm}

\begin{figure}[htb]
\begin{center}
\includegraphics[width=0.99\linewidth]{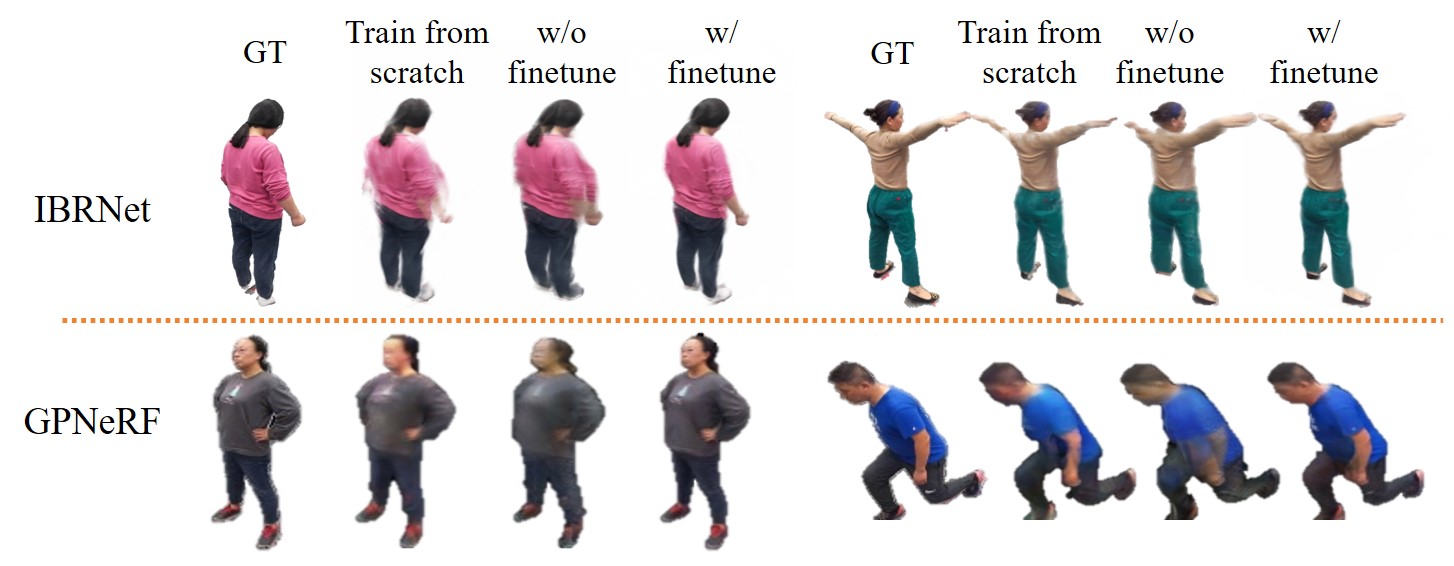}
\end{center}
\vspace{-3mm}
\caption{\textbf{Qualitative comparison of IBRNet and GPNeRF on the test set of HuMMan.} Without finetuning, the models only trained on \thename++ may suffer from domain gap. With some time for finetuning, the models outperform the ones trained merely on the train set of HuMMan.}
\label{fig:nerf_vis_humman}
\end{figure}

\subsection{Per-Subject 3DGS Reconstruction for Human}\label{per_subject_gs}
\begin{figure*}
\begin{center}
\includegraphics[width=1.0\textwidth]{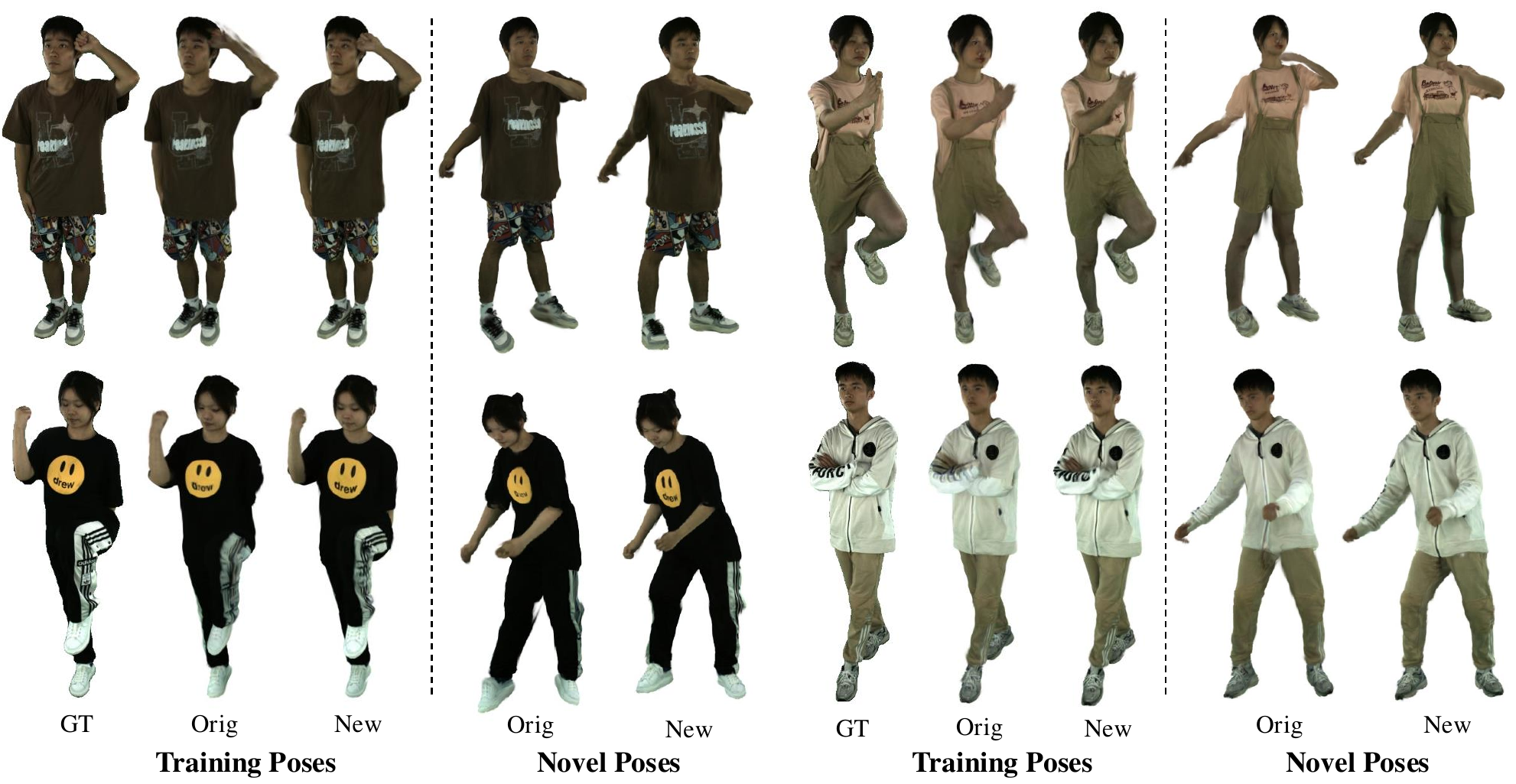}
\end{center}
\vspace{-2mm}
\caption{
\textbf{The visualization results of Animatable Gaussians for both novel view and novel pose synthesis.} \textbf{GT} denotes ground truth, \textbf{Orig} refers to the original version of SMPLX, and 
\textbf{New} refers to the updated SMPLX from \thename++.}
\label{fig:animate_gaussian_result}
\end{figure*}

\begin{figure}[htbp]
\begin{center}
\includegraphics[width=0.99\linewidth]{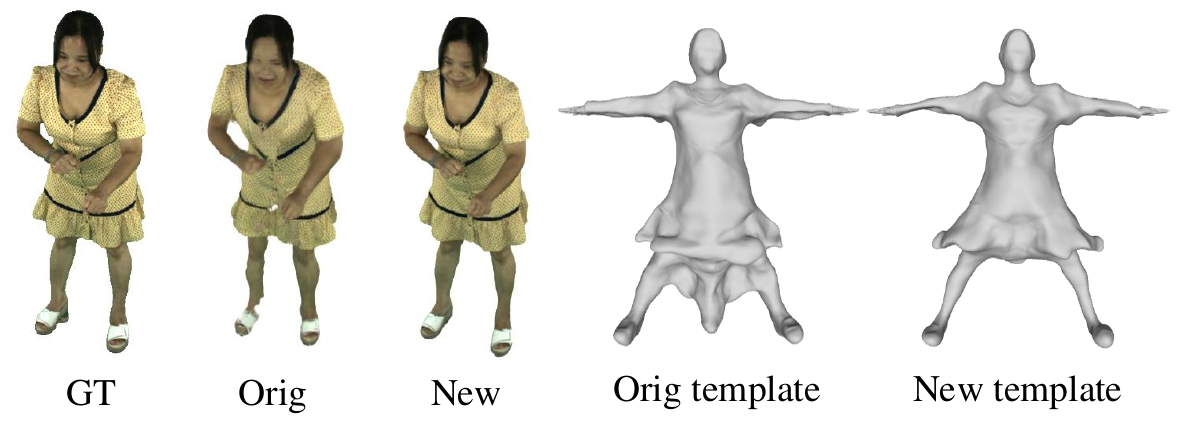}
\end{center}
\vspace{-2mm}
\caption{
The visualization results of Animatable Gaussians for loose-fitting novel view synthesis and the corresponding parametric template.}
\label{fig:animate_gaussian_dress_result}
\end{figure}

Recently, 3D Gaussian splatting~\cite{kerbl20233d}, characterized by its explicit neural representation and remarkable rendering quality, has emerged as a promising alternative to NeRFs. Building on this advancement, Animatable Gaussians~\cite{li2024animatable} introduces a novel avatar representation that leverages 3D Gaussian splatting and powerful 2D CNNs to achieve realistic avatar modeling from multi-view human images. Thus, we adopt Animatable Gaussians as the baseline method to validate the effectiveness of high-quality SMPLX annotations. For this purpose, we use 16 views—randomly selecting 12 views for optimizing Gaussian parameters and reserving the remaining 4 views for evaluation. We randomly choose 20 subjects to conduct experiments and compute the results averaged across these subjects. The quantitative results are shown in Tab.~\ref{tab:animate_gaussian_metric}, where both novel view and novel pose synthesis achieve more realistic reconstruction results using the new SMPLX parameters estimated via the advanced approach. These results also indicate that \thename++ can better support learning-based reconstruction methods in the task of per-subject animatable human reconstruction. We provide qualitative results in Fig.~\ref{fig:animate_gaussian_result} to visualize the differences in reconstruction quality. For loose-fitting clothing, the reconstructed template and rendering results are also visualized in Fig.~\ref{fig:animate_gaussian_dress_result}.

\begin{table}[htbp]
    % \resizebox{\columnwidth}{!}{
    \centering
    \setlength{\tabcolsep}{10pt} % 增加列间距
    {
    
    \begin{tabular}{c|ccc}
        \toprule
        SMPLX-Annotation  & PSNR $\uparrow$ & SSIM $\uparrow$ & LPIPS $\downarrow$ \\
        \midrule
        Orig SMPLX & 27.259 & 0.968 & 0.0452 \\
        New SMPLX & 28.593 & 0.976 & 0.0369 \\

        % Ours &\cellcolor{deeppeach}27.568 & \cellcolor{deeppeach}0.954 & \cellcolor{lightgray} 0.0682 \\
        \bottomrule
    \end{tabular}
    }
    \caption{
    % Performance of different models. 
    Quantitative evaluation of Animatable Gaussians on the \thename++ dataset using different versions of SMPLX annotations.
    }
    \vspace{-1mm}
    \label{tab:animate_gaussian_metric}
\end{table}

\subsection{Generalizable 3DGS Reconstruction for Human}\label{general_gs}
\begin{figure}[ht]
\begin{center}
\includegraphics[width=0.99\linewidth]{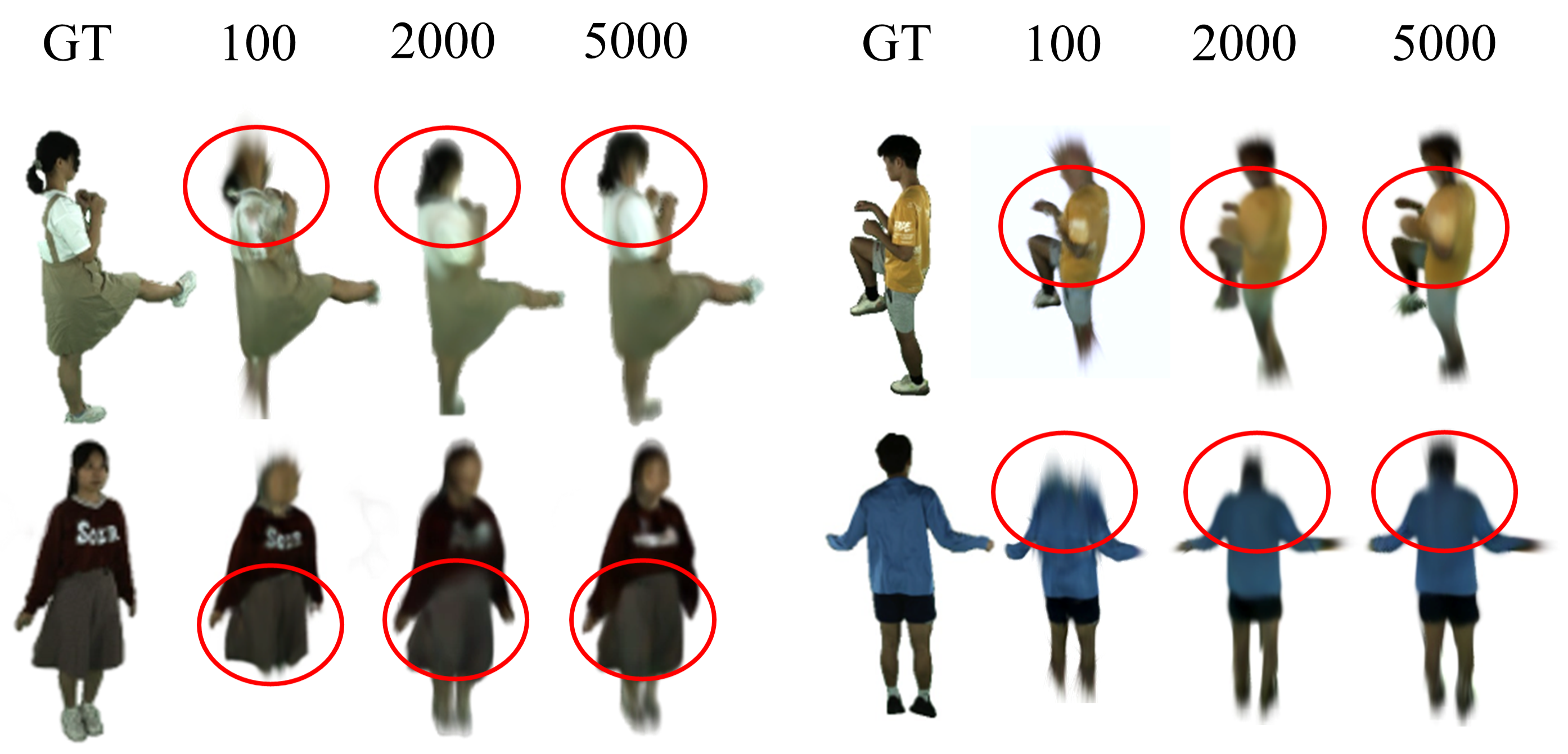}
\end{center}
\vspace{-3mm}
\caption{
\textbf{Novel view synthesis results of LaRa on the test data of \thename++.} \textbf{GT} means ground truth. The number of \textbf{100}, \textbf{2000}, and \textbf{5000} indicate the respective quantities of outfits utilized during the training process.
}
\vspace{-2mm}
\label{fig:lara_results}
\end{figure}

\begin{figure*}
\begin{center}
\includegraphics[width=1.0\textwidth]{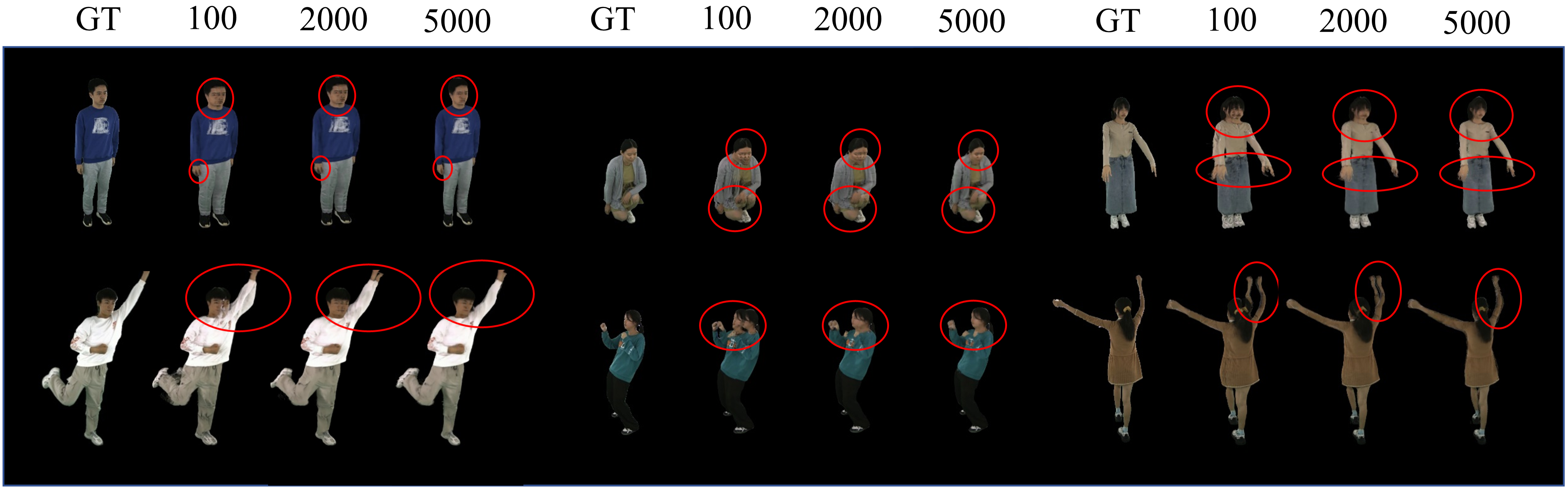}
\end{center}
\vspace{-2mm}
\caption{
\textbf{Novel view synthesis results of EVA-Gaussian trained on \thename++.}  \textbf{GT} means ground truth. The number of \textbf{100}, \textbf{2000}, and \textbf{5000} indicate the respective quantities of outfits utilized during the training process.}
\label{fig:eva_gaussian_results}
\end{figure*}

Feed-forward 3D Gaussian Splatting method has demonstrated exceptional capability and  achieved fast reconstruction in novel view synthesis compared with optimization based methods. To explore the applicability of \thename++ to generalizable multi-view human reconstruction, we conduct experiments on LaRa~\cite{chen2024lara}  which represents scenes as Gaussian Volumes for large-baseline radiance field reconstruction and EVA-Gaussian~\cite{hu2024eva}  leverages hybrid multi-stage feature encoding to achieve high-quality generalizable reconstructions. To systematically evaluate the impact of training data scale, we train LaRa and EVA-Gaussian both from scratch and with varying numbers of identity samples: 100, 2000, and 5000. In our experiment, LaRa utilizes four evenly distributed views as input across 360 degrees, and infers the novel view results, while EVA-Gaussian uses two views, with the angle between views being 45 degrees.
For EVA-Gaussian, we pretrain a depth estimator using rendered human depth maps as ground truth in the first stage, which is then used for Gaussian parameter prediction in the second stage.
The quantitative comparisons of the results are presented in Tab.~\ref{tab:gene_human_gs} while the visualization results can be found in Fig.~\ref{fig:lara_results} and Fig.~\ref{fig:eva_gaussian_results}.
Experimental results demonstrate that as the number of training identities increases, the rendered novel-view images from both methods exhibit more robust Gaussian point localization and improved rendering quality, indicating enhanced generalization ability in human reconstruction. Benefiting from the depth map predictions of the Efficient Cross-View Attention (EVA) module, EVA-Gaussian achieves satisfactory results, while LaRa is limited by its Gaussian volume representation and the larger baseline of the input human images. Please note that we use the training version of EVA-Gaussian without the anchor loss, which requires additional data processing and landmark generation during the training stage.

% Due to the misalignment in depth ranges between THuman2.0~\cite{yu2021function4d} (the dataset used for training the original EVA-Gaussian) and MVHumanNet, we train EVA-Gaussian from scratch and fine-tune the official LaRa checkpoint on MVHumanNet.

%  Benefiting from our dataset, which includes depth maps rendered from 2DGS optimization for each subject, we first train a depth estimator using the rendered depth maps in the first stage of EVA-Gaussian. In the second stage, Gaussian attribute estimation is performed by loading the estimated depth from the first stage and further predicting other Gaussian parameters.

% To systematically evaluate the impact of training data scale, we train both stages from scratch using varying numbers of identity samples: 100, 2000, and 5000. For all settings, we maintain consistent hyperparameters, including the same learning rate and the same number of training steps for each stage. The evaluation set remains identical across experiments to ensure a fair comparison. As shown in \cref{fig:eva_gaussian_result}, with an increasing number of training identities, the rendered novel view images exhibit more robust Gaussian point localization and improved rendering quality, demonstrating enhanced generalization ability in human reconstruction. Quantitive results are shown in \cref{tab:evagaussian_metric}.

\begin{table}[tb]
    \resizebox{\columnwidth}{!}{
    % \large
    \begin{tabular}{c|ccc|ccc}
        \toprule
         %&  &  \multicolumn{3}{c}{\makecell{Static Scenes \\ w/ Ground Truth}} & \multicolumn{3}{c}{\makecell{Dynamic Scenes \\ w/o Ground Truth}} \\
        {\multirow{2}*{\begin{tabular}[c]{@{}c@{}}Number of \\ outfits\end{tabular}}}   &  \multicolumn{3}{c|}{LaRa~\cite{chen2024lara} }   & \multicolumn{3}{c}{EVA-Gaussian~\cite{hu2024eva}}   \\
         &  PSNR $\uparrow$ &  SSIM $\uparrow$ &  LPIPS $\downarrow$ &  PSNR  $\uparrow$ &  SSIM $\uparrow$ &  LPIPS $\downarrow$ \\
        \midrule
             100     & 21.431 & 0.935 & 0.0840   &  26.984 & 0.952 & 0.0515 \\
            2000    & 21.869 & 0.937 & 0.0846 & 27.363 & 0.960 & 0.0465 \\
            5000    & 22.441 & 0.941 & 0.0793  & 28.544 & 0.968 & 0.0401 \\
        \bottomrule
    \end{tabular}
    }
    \vspace{-1mm}
    \caption{Quantitative comparison of LaRa and EVA-Gaussian with different scales of data for training.}
    \vspace{-2mm}
     % \vspace{-4mm}
    \label{tab:gene_human_gs}
\end{table}

% \begin{table}[htbp]

%     \centering
%     \setlength{\tabcolsep}{10pt} % 增加列间距
%     {
%     \begin{tabular}{c|ccc}
%         \toprule
%         Number of outfits  & PSNR $\uparrow$ & SSIM $\uparrow$ & LPIPS $\downarrow$ \\
%         \midrule
%         w/o Finetune & 19.247 &  0.934 &  0.0807 \\
%         100 & 21.431 & 0.935 & 0.0840 \\
%         2000 & 21.869 & 0.937 & 0.0846 \\
%         5000 & 22.441 & 0.941 & 0.0793 \\
%         % Ours &\cellcolor{deeppeach}27.568 & \cellcolor{deeppeach}0.954 & \cellcolor{lightgray} 0.0682 \\
%         \bottomrule
%     \end{tabular}
%     }
%     \caption{
%     % Performance of different models. 
%     Quantitative evaluation of LaRa's rendering qualit on the MVHumanNet dataset.
%     }
%     \label{lara_metric}
% \end{table}

% \begin{table}[htbp]
%     % \resizebox{\columnwidth}{!}{
%     \centering
%     \setlength{\tabcolsep}{10pt} % 增加列间距
%     {
    
%     \begin{tabular}{c|ccc}
%         \toprule
%         Number of outfits  & PSNR $\uparrow$ & SSIM $\uparrow$ & LPIPS $\downarrow$ \\
%         \midrule
%         100 & 26.984 & 0.952 & 0.0515 \\
%         2000 & 28.363 & 0.960 & 0.0465 \\
%         5000 & 29.544 & 0.968 & 0.0401 \\
%         % Ours &\cellcolor{deeppeach}27.568 & \cellcolor{deeppeach}0.954 & \cellcolor{lightgray} 0.0682 \\
%         \bottomrule
%     \end{tabular}
%     }
%     \caption{
%     % Performance of different models. 
%     Quantitative evaluation of EVA-Gaussian's rendering quality on the MVHumanNet dataset.
%     }
%     \label{tab:eva_gaussian_metric}
% \end{table}

\subsection{Text-driven Image Generation}\label{text_task}
\begin{figure}[tb]
\begin{center}
\includegraphics[width=1.0\linewidth]{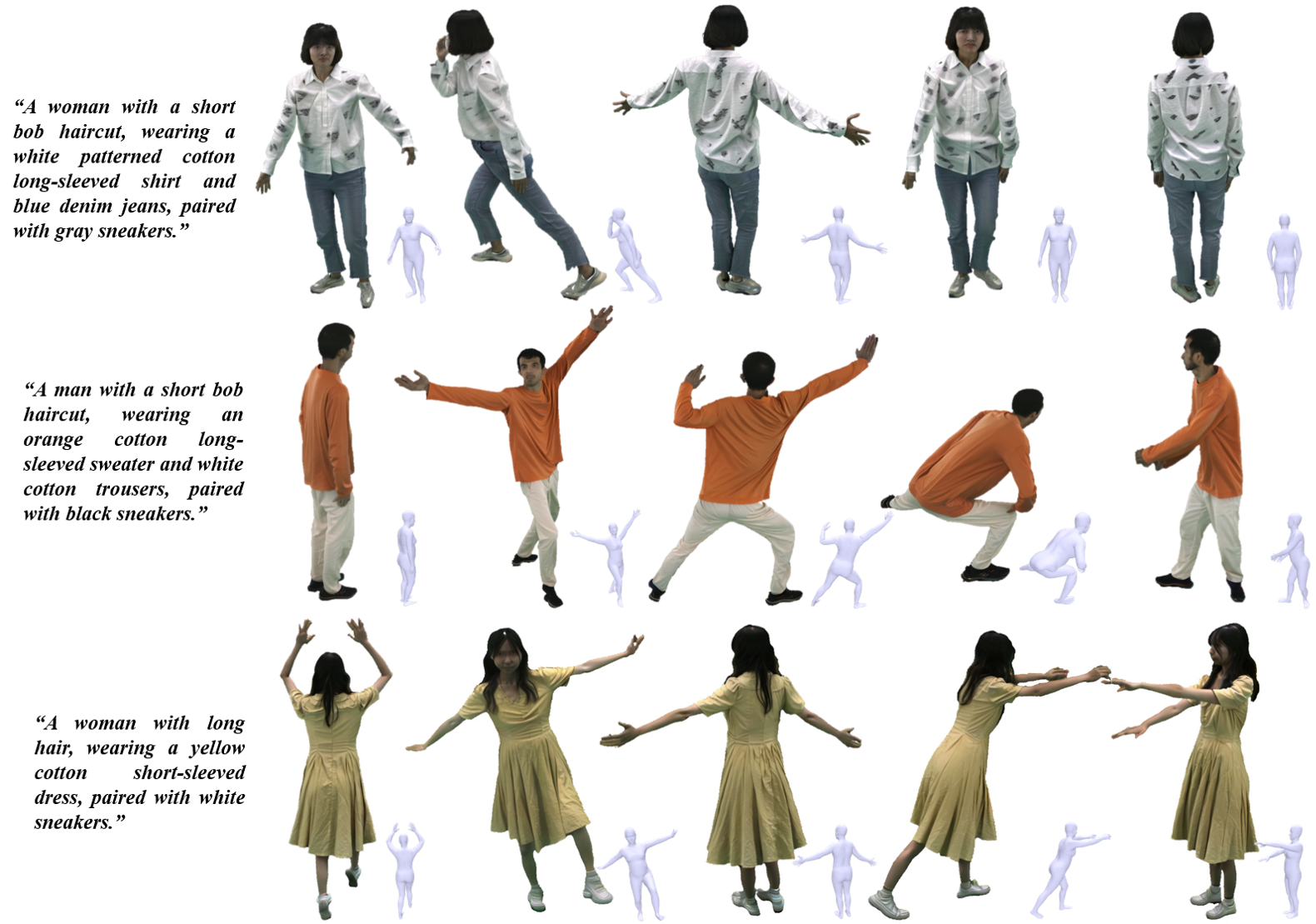}
\end{center}
%\vspace{-1mm}
\vspace{-3mm}
\caption{
\textbf{The visualization of images generated by text-to-image model trained on \thename++ with SMPL condition and text prompts as input.} The results demonstrate that training on our large-scale high-quality human dataset enables the generation of high-resolution human images using textual description and SMPL conditions.}
\vspace{-5mm}
\label{fig:text_driven image generation}
\end{figure}

\thename++~is able to serves as a fundamental resource for our text-driven image generation method. The inclusion of comprehensive pose variations within our dataset enhances the potential for generating diverse human images in accordance with text descriptions. 
%This valuable dataset greatly facilitates the creation of a wide range of human images based on textual prompts.
%As mentioned above, we collect a large-scale dataset of multi-view human images along with detailed textual descriptions, which provide essential support for our text-driven image generation. In addition, the richness of pose variations within our dataset expands the possibilities for our generation, significantly facilitating the creation of diverse human images based on textual descriptions.
We finetune the powerful text-to-image model, Stable Diffusion~\cite{rombach2022high} on \thename++ dataset to enable text-driven realistic human image generation. 
As shown in Fig.~\ref{fig:text_driven image generation}, given a text description and a target SMPL pose, we can produce high-quality results with the same consistency as text description and SMPL.

% %\qlt{TODO: hongge write motivation to multiview-consistency novel view synthesis}
% In addition, leveraging the multi-view nature of the human data with text annotation provides essential support for our text-driven with camera controllable novel-view human generation. Following the concurrent novel-view synthesis approaches~\cite{liu2023zero,shi2023mvdream}, we train a multi-view diffusion model on MVHumanNet that conditioned on camera-pose and SMPL parameters, to generate view-consistent human novel-view synthesis.  As shown in Fig.~\ref{fig:text_driven multiview generation}, given a textual description and  SMPL parameters with a camera pose sequence  as input, we can generate view-consistent human images aligned with the corresponding SMPL models.   

Based on the results derived from the text-driven image generation, it becomes evident that the utilization of large-scale multi-view data from real capture contributes to the efficacy of text-driven realistic human image generation.

\subsection{Human Generative Model }\label{avatar_task}

Recently, generative models have become a prominent and highly researched area. 
Methods such as StyleGAN~\cite{karras2020analyzing, fu2022styleganhuman} have emerged as leading approaches for generating 2D digital human. 
More recently, the introduction of GET3D~\cite{gao2022get3d} has expanded this research area to encompass the realm of 3D generation. 
%With the availability of substantial data in MVhumanNet, we embark on an exploratory journey as pioneers, aiming to investigate the potential applications of existing 2D and 3D generative models by leveraging a large-scale dataset comprising real-world multi-view full-body data. 
With the availability of massive data in \thename++, we embark on an exploratory journey as pioneers, aiming to investigate the potential applications of existing 2D and 3D generative models by leveraging a large-scale dataset comprising real-world multi-view full-body data. 
We conduct experiments to unravel the possibilities within this context.

\begin{figure}[tb]
\begin{center}
\includegraphics[width=0.96\linewidth]{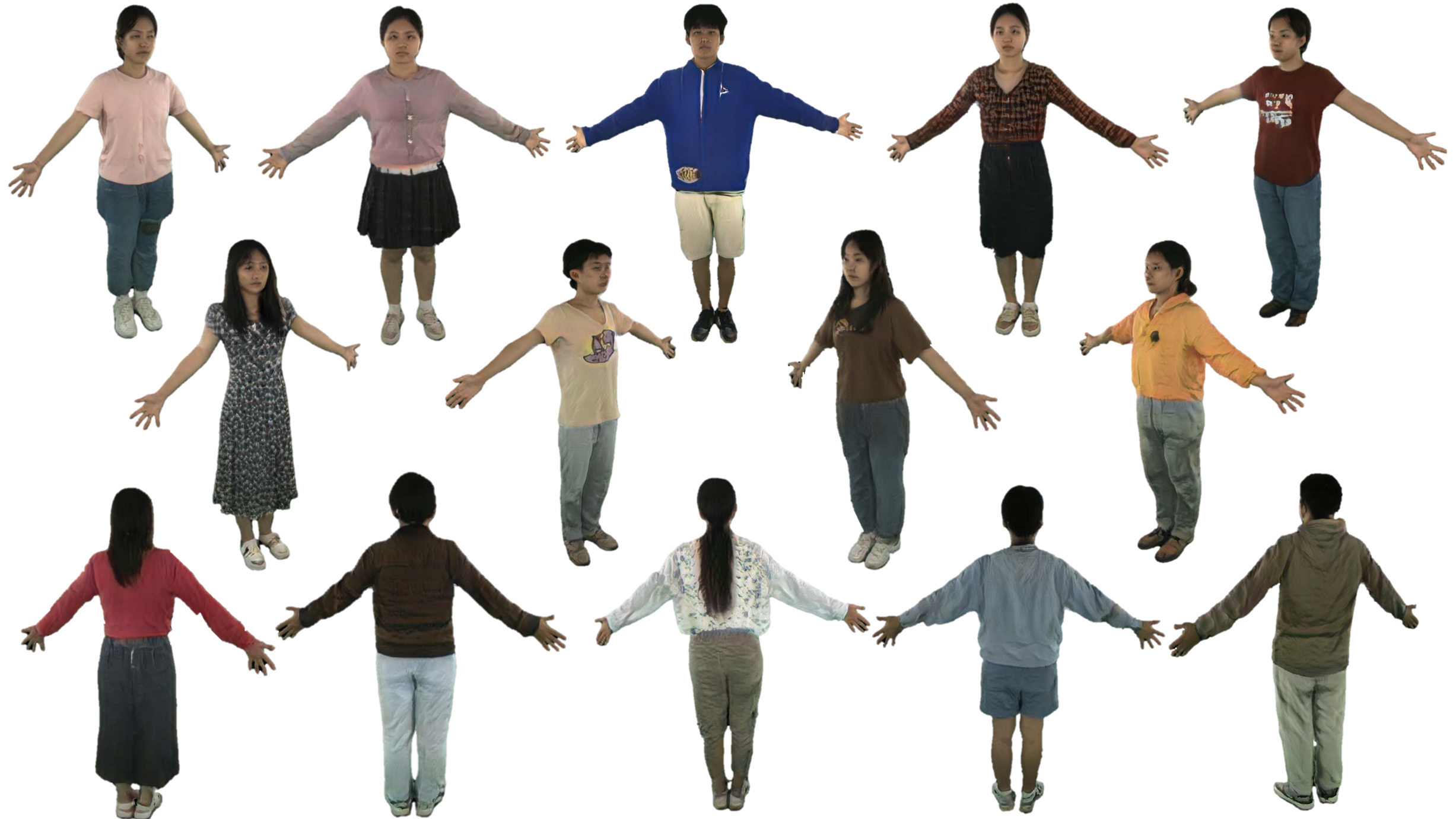}
\end{center}
\vspace{-3mm}
\caption{
\textbf{Visualize the results of StyleGAN2 trained with \thename++.} We randomly sample latent codes from Gaussian distribution and obtain the results.}
\label{fig:stylegan2_human}
\end{figure}

% \vspace{-3mm}
\noindent\textbf{2D Generative Model} Giving a latent code sampled from Guassian distribution, StyleGAN2 outputs a reasonable 2D images. 
In this part, we feed approximately 198,000 multi-view A-pose images (5500 outfits) and crop to 1024$\times$1024 resolution into the network with camera conditions for training. Fig.~\ref{fig:stylegan2_human}
visualizes the results. 
Our model not only produces frontage full-body images but also demonstrates the capability to generate results from other views, including the back and side views.

%\begin{figure}[tb]
%\begin{center}
%\includegraphics[width=1.0\linewidth]{figure/qlt_text.png}
%\end{center}
%\vspace{-6mm}
%\captionsetup{skip=10pt}
%\caption{
%\textbf{Example images generated by multi-view diffusion model trained on MVHumanNet with SMPL condition and input test prompts.} The results clearly show that training multi-view diffusion model on our large-scale multi-view human dataset will get satisfactory view-consistent result with the same subjects.}
%\label{fig:text_driven multiview generation}
%\vspace{-4mm}
%\end{figure}

\begin{figure}[tb]
\begin{center}
\includegraphics[width=0.97\linewidth]{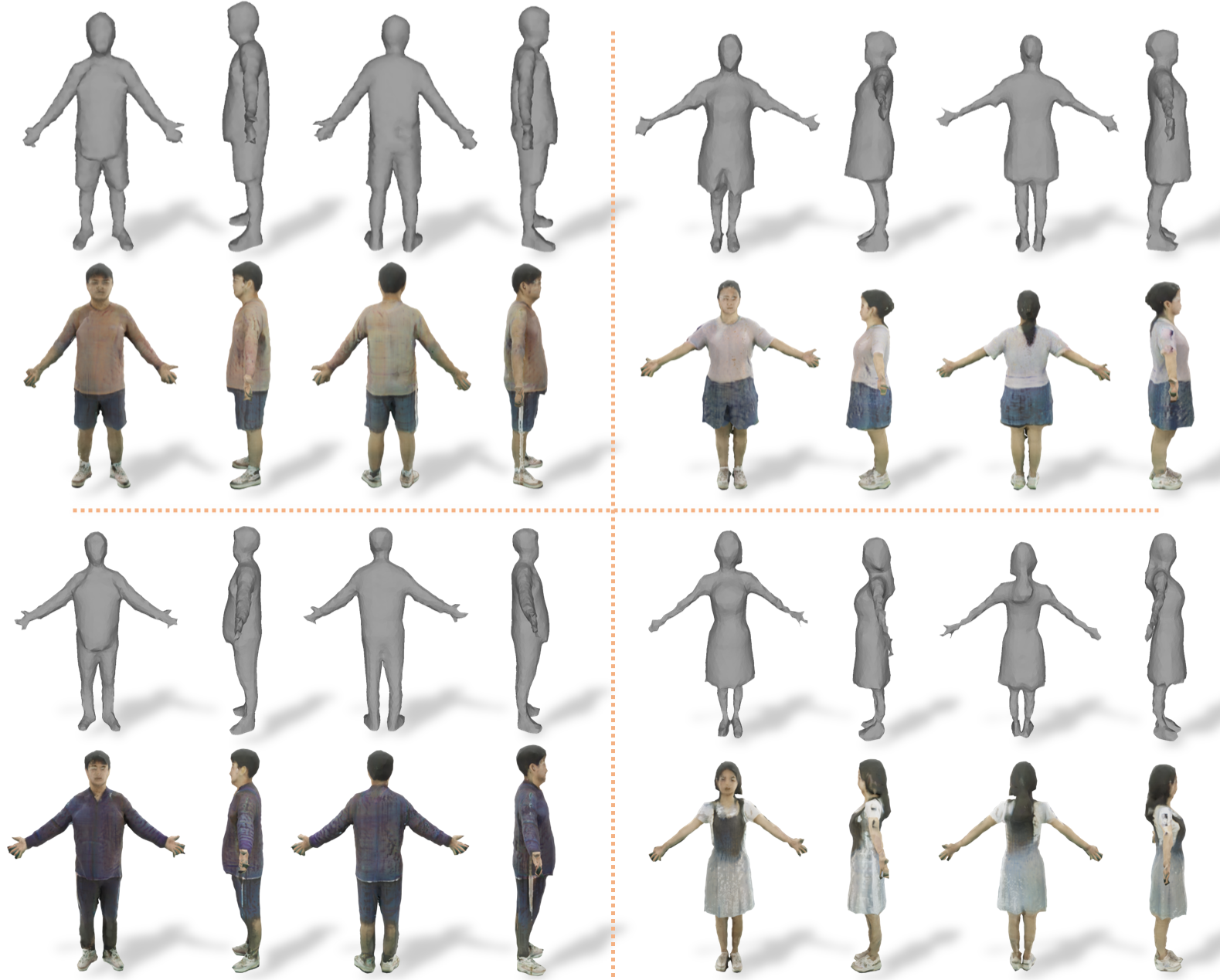}
\end{center}
\vspace{-3mm}
\caption{
\textbf{The visualization results of GET3D trained with \thename++ rendered by Blender~\cite{blender}.} The first and third rows represent the geometry, while the second and fourth row shows the texture corresponding to geometry. }
\vspace{-3mm}
\label{fig:get3d}
% \vspace{-3mm}
\end{figure}

\noindent\textbf{3D Generative Model}
Unlike StyleGAN2, GET3D~\cite{gao2022get3d} introduces a distinct requirement of one latent code for geometry and another for texture. 
%The network leverages DMTet~\cite{shen2021dmtet} to generate the geometry, which is subsequently colorized to produce a 3D mesh with texture. 
%During the training phase of the network, geometry generation is constrained by masks derived from multi-view data, while texture field generation is constrained by images. 
We use the same amount of data as training StyleGAN2 to train GET3D. 
The visualization results are shown in Fig.~\ref{fig:get3d}. 
The model exhibits the ability to generate reasonable geometry and texture in the A-pose, thereby enabling its application in various downstream tasks.
With the substantial support provided by \thename++, various fields, including 3D human generation, can embark on further exploration by transitioning from the use of synthetic data or single-view images to the incorporation of authentic multi-view data.
We also conduct experiments to prove that the performance of the generative model will become more powerful with the increase in the amount of data. The quantitative results are shown in Tab.~\ref{tab:num_generative}. 
We have reason to believe that with the further increase of data, the ability of trained models can further improve.

\begin{table}[tb]
    \centering
    % \scalebox{0.8}
    {%\def\arraystretch{1} \tabcolsep=0.2em 
    \begin{tabular}{c|cc}
        \toprule
          {\multirow{2}*{Number of Subjects}}  &  \multicolumn{2}{c}{FID$\downarrow$} \\
          & StyleGAN2~\cite{karras2020analyzing}  &  GET3D~\cite{gao2022get3d} \\
        \midrule
          3000 &  14.05 &  41.54   \\
          5500 &  7.08 \textcolor{blue}{(-6.97)} &  25.12 \textcolor{blue}{(-16.42)}\\  
        \bottomrule
    \end{tabular}
    }
    \captionsetup{skip=6pt}
    \vspace{-1mm}
    \caption{\small \textbf{Quantitative comparison of generative models with different data scale}. 
    The performance of both 2D and 3D generative models exhibits obvious improvement with scaling up data.} 
    \vspace{-4mm}
    %\captionsetup{skip=4pt}
    \label{tab:num_generative}
    
\end{table}

\begin{figure}[htbp]
\begin{center}
\includegraphics[width=0.95\linewidth]{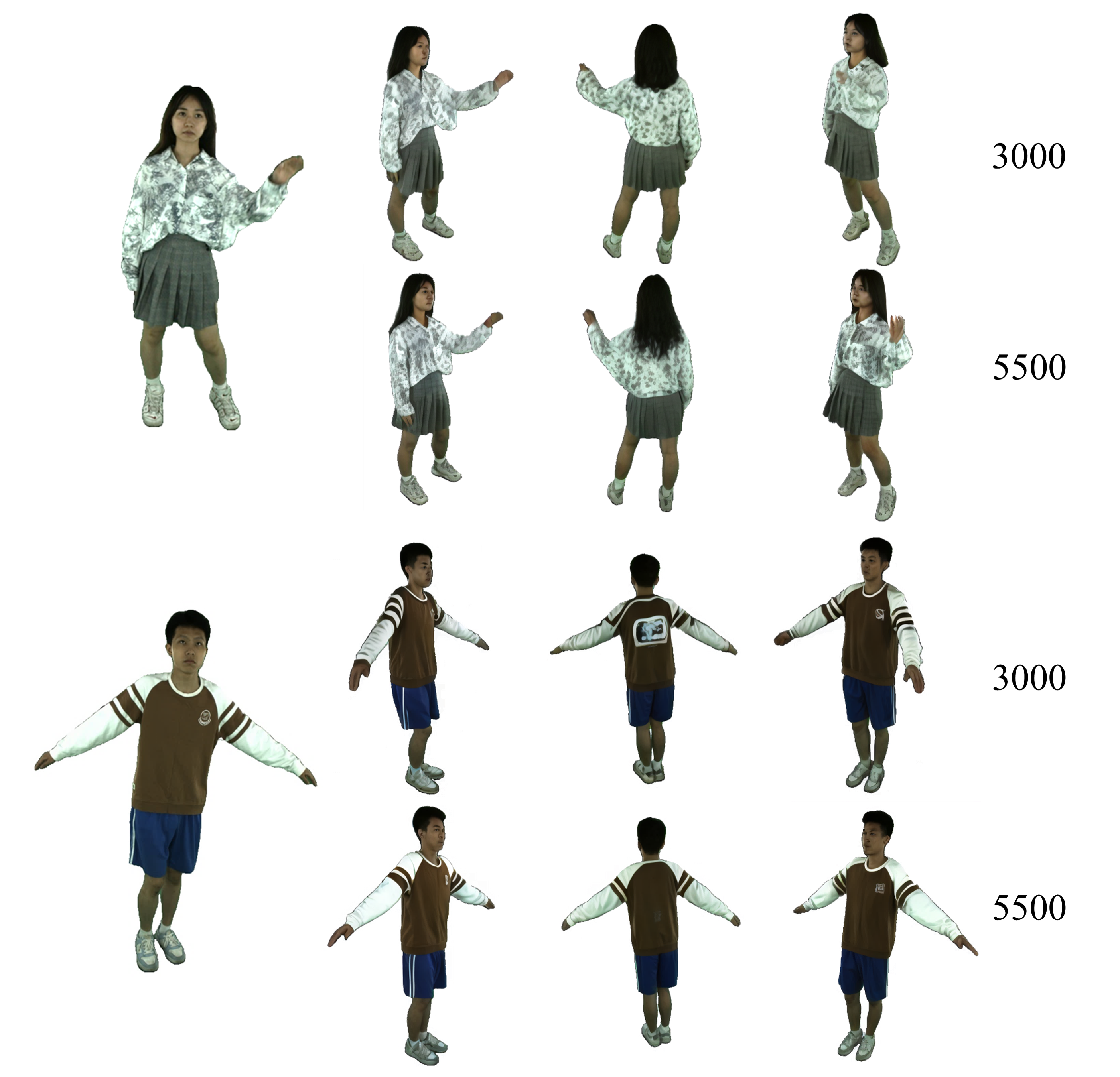}
\end{center}
\vspace{-3mm}
\caption{
\textbf{The visualization results of MVdream fine-tuned on \thename++.} As the scale of training data increases, the multi-view generation results become more reasonable, particularly for back-view hair and texture, as well as side-view poses.}

\label{fig:multiview_result}
\end{figure}

\noindent\textbf{Multi-view Generative Model} For multi-view generation, Zero-1-to-3 pioneers open-world single-image-to-3D conversion through zero-shot novel view synthesis. We use the same amount of data as required for training 2D and 3D generative models, cropping images to a resolution of 512×512 and integrating them into the Stable Diffusion v2.1 base model of MVDream~\cite{shi2023mvdream}. We also conduct experiments to demonstrate that the performance of the generative model improves as the amount of data increases. The quantitative results are presented in Tab.~\ref{tab:multi_view_gen_metric}, and the visualization results are shown in Fig.~\ref{fig:multiview_result}. We observe that under the latent diffusion setting, facial results are particularly sensitive and may suffer from distortion, as these areas occupy only a small portion of the overall pixel. However, we have reason to believe that with further data increases and improvements in method design, the capabilities of trained models can improve even further. 

\begin{table}[htbp]
    % \resizebox{\columnwidth}{!}{
    \centering
    \setlength{\tabcolsep}{10pt} % 增加列间距
    {
    
    \begin{tabular}{c|ccc}
        \toprule
        Number of subjects  & PSNR $\uparrow$ & SSIM $\uparrow$ & LPIPS $\downarrow$ \\
        \midrule
        3000 & 16.811 & 0.924 & 0.171 \\
        5500 & 18.328 & 0.934 & 0.146 \\

        % Ours &\cellcolor{deeppeach}27.568 & \cellcolor{deeppeach}0.954 & \cellcolor{lightgray} 0.0682 \\
        \bottomrule
    \end{tabular}
    }
    \vspace{-1mm}
    \caption{
    % Performance of different models. 
    Quantitative evaluation of multi-view human generative models with different data scale.
    }
    \vspace{-6mm}
    \label{tab:multi_view_gen_metric}
\end{table}

\begin{table}[tb]
    % \resizebox{\columnwidth}{!}{
    \centering
    \setlength{\tabcolsep}{10pt} % 增加列间距
    {
    
    \begin{tabular}{c|cc}
        \toprule
        Number of subjects  & Rel $\downarrow$ & $\tau (thresh=1.03) \uparrow$  \\
        \midrule
        3000 & 5.356 & 35.855  \\
        5500 & 3.857 & 52.189  \\

        % Ours &\cellcolor{deeppeach}27.568 & \cellcolor{deeppeach}0.954 & \cellcolor{lightgray} 0.0682 \\
        \bottomrule
    \end{tabular}
    }
    \vspace{-1mm}
    \caption{
    % Performance of different models. 
    Quantitative evaluation of fine-tuning DUSt3R with different data scale. 
    }
    \label{tab:dust3r_metric}
\end{table}

\subsection{Reconstruction from Unconstraint Human Images}\label{dust3r_task}
\begin{figure}[htbp]
\begin{center}
\includegraphics[width=0.95\linewidth]{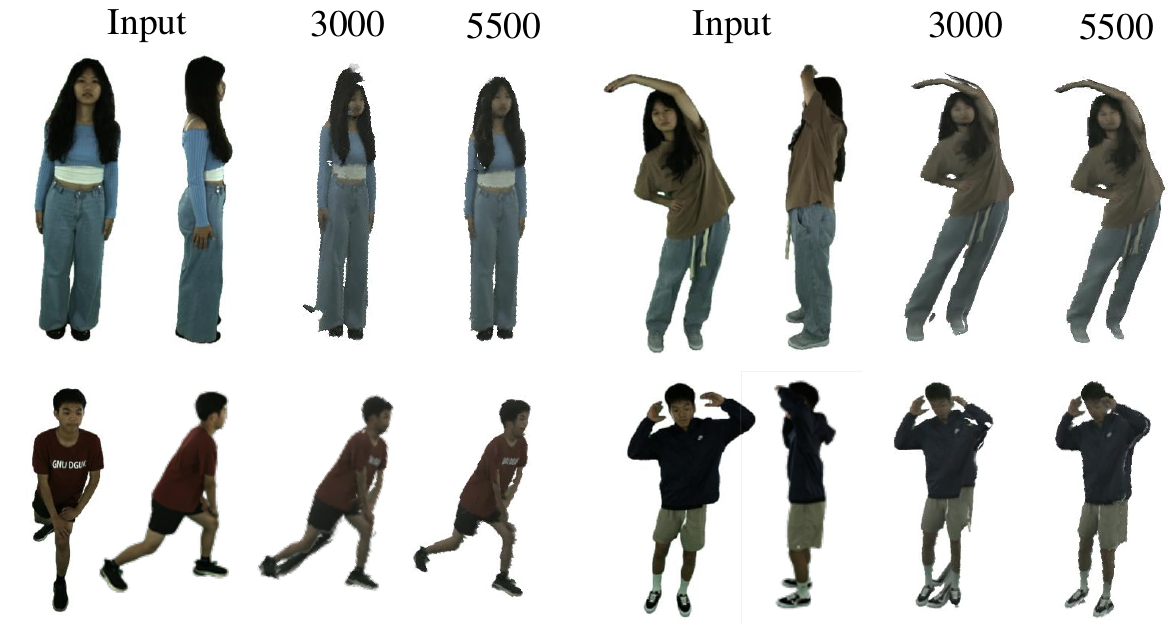}
\end{center}
\vspace{-3mm}
\caption{
\textbf{The visualization results of DUSt3R fine-tuned on \thename++.} We  visualize the colored point cloud from two input images.}
\vspace{-3mm}
\label{fig:dust3r}
\end{figure}

3D human reconstruction from unconstrained images poses a significant challenge in computer vision, primarily due to the complexities associated with pose estimation and shape recovery. Recently, DUSt3R~\cite{wang2024dust3r} introduced an innovative approach to address this challenge by predicting point maps for a pair of uncalibrated stereo images in a unified coordinate system with implicit correspondence searching. However, since DUSt3R is not trained on human-centric datasets, the results on human data are unsatisfactory. Therefore, we fine-tune DUSt3R on the \thename++ dataset and conduct experiments to demonstrate that the model's performance significantly improves with the expansion of the training data scale. The quantitative results are presented in Tab.~\ref{tab:dust3r_metric}, and the visualization results are shown in Fig.~\ref{fig:dust3r}. From the experimental results, we observe that as the scale of the dataset increases, the depth ambiguity of the point map generated by DUSt3R is significantly reduced, thereby enhancing overall performance.

\section{Conclusions}
In this work, we present \thename++, a large-scale multi-view dataset containing \textbf{4,500} human identities, \textbf{9,000} daily outfits and \textbf{645 million} frames with extensive annotations. Additionally, the proposed \thename++ dataset is enhanced with newly processed normal maps and depth maps, significantly expanding its applicability and utility for advanced human-centric research. We primarily focus on the domain of collecting daily dressing, which allows us to easily scale up the human data. To probe the potential of the proposed large-scale dataset, we design various experiments to demonstrate how \thename++ can be utilized to advance these 3D human reconstruction tasks, including some of the latest methods in the field. We plan to release the \thename++ dataset with annotations publicly and hope that it will serve as a foundation for further research in the 3D digital human community.

\clearpage

\small
\bibliographystyle{IEEEtran}
\bibliography{IEEEabrv,main}

\end{document}